\documentclass[a4paper, 10pt, conference]{ieeeconf}      
\IEEEoverridecommandlockouts                              
\usepackage{graphicx} 
\usepackage[english]{babel}
\usepackage{epsfig} 
\usepackage{wrapfig}
\usepackage{times} 
\usepackage{mathptmx} 
\usepackage{amsmath} 
\usepackage{amssymb}  
\usepackage[figuresright]{rotating}
\usepackage{float}
\usepackage{subfig}
\usepackage{tabularx}
\usepackage{rotating}
\usepackage{url}

\title{\LARGE \bf Heterogeneity for Increasing Performance and Reliability of Self-Reconfigurable Multi-Robot Organisms}

\author{S.~Kernbach$^1$, F.~Schlachter$^1$, R.~Humza$^{2}$, J.~Liedke$^4$, S.~Popesku$^1$, S.~Russo$^3$, T.~Ranzani$^3$, L.~Manfredi$^3$,\\ C.~Stefanini$^3$, R.~Matthias$^4$, Ch.~Schwarzer$^6$, B.~Girault$^5$, P.~Alschbach$^1$, E.~Meister$^1$, O.~Scholz$^2$\\[2mm]
\small
$^1$Institute of Parallel and Distributed Systems, University of Stuttgart, Germany,
$^2$Fraunhofer Institute for Biomedical \\Engineering (IBMT), St. Ingbert, Germany,
$^3$The BioRobotics Institute, Scuola Superiore Sant'Anna, Italy,\\
$^4$Institute for Process Control and Robotics, Karlsruhe Institute of Technology (KIT), Germany,
$^5$\'Ecole Normale \\Supérieure de Cachan, Cachan, France, 
$^6$Institute for Evolution and Ecology, University of Tuebingen, Germany\\
\thanks{Contact author: serge.kernbach@ipvs.uni-stuttgart.de. IROS11, workshop on ``Reconfigurable Modular Robotics", San Francisco, 2011.}
\vspace{-5mm}}

\begin{document}

\maketitle
\thispagestyle{empty}
\pagestyle{empty}

\begin{abstract}
Homogeneity and heterogeneity represent a well-known trade-off in the design of modular robot systems. This work addresses the heterogeneity concept, its rationales, design choices and performance evaluation. We introduce challenges for self-reconfigurable systems, show advances of mechatronic and software design of heterogeneous platforms and discuss experiments, which intend to demonstrate usability and performance of this system.
\end{abstract}

\section{Introduction}

Modular and reconfigurable robotics represents the area of mechatronic systems with dynamically changing structures and functionalities~\cite{Ostergaard:2006}. The principle of modularity is useful in obtaining several advanced properties such as reliability, adaptation~\cite{Kernbach08b},  encapsulation of complexity~\cite{Kernbach08_2}, or in exploiting evolutionary and self-developmental capabilities of artificial systems~\cite{Kernbach09_CEC}. Not only technological but also natural complex biological systems utilize the concept of modularity, which is known as multicellularity. Currently, artificial multicellularity attracted attention of researchers in exploring new biological inspirations, homeostatic mechanisms, macroscopic regulation and other issues~\cite{Levi10}.

State of the art solutions in both lattice-based and chain-based reconfigurable systems count primarily homogeneous platforms, e.g.~\cite{salemi2006sdm}, \cite{stoy06}, \cite{christensen04}, \cite{murata2001csr}. There are multiple reasons for such a development: complexity of mechanic, electronic and software design, exploration of reconfiguration approaches, finding a balance between "rigid" and "soft" design principles. In this work we primarily focus on challenges and driving forces for introducing heterogeneity into modular and reconfigurable systems.

Homogeneity and heterogeneity provide different advantages and represent two opposite points on the scale of universality and specialization~\cite{Kernbach09Nep}. For instance, homogeneous elements can be easily replaced, such systems are more scalable. Most artificial and natural swarms consist of structurally homogeneous elements, capable of behavioral specialization, e.g.~\cite{Kornienko_S06b}. However, increasing the complexity of the system, e.g. by aggregation into multi-robots organisms, we encountered bottlenecks in complexity of homogeneous elements, which are expressed in terms of weight, power limitation, or locomotion capabilities. 

This paper addresses the research and technological challenge of finding the compromise between homogeneous and heterogeneous design principles in general, and in particular within the scope of the European funded projects \textit{Replicator} and \textit{Symbrion}. First of all, the degree of heterogeneity is an open-ended trade-off between complexity, scalability, cost factors, and required functionalities~\cite{Kernbach10SoSA}. Finding a compromise, especially in the initial stage of development, is a tough problem~\cite{Kernbach08}. To establish a rationale for the heterogeneous design, so-called challenges for modularity and reconfiguration are introduced. These challenges define not only the design choices but also set up experiments for performance measurement. We discuss these issues in Sec.~\ref{sec:challenges}. 

Heterogeneity requires specific technological solutions for docking elements, energy and communication buses, mechatronics, power management and others. The developed solutions are shortly overviewed in Sec.~\ref{sec:solutions}. Special attention is paid to compatibility and common elements of this system as well as to a specialization of corresponding platforms. Secs.~\ref{sec:manufacturing},   \ref{sec:software} and Sec.~\ref{sec:experiments} introduce a software architecture, manufacturing issues and two experiments, which aim at a qualitative demonstration of increased performance and reliability of this system. Finally, in Sec.~\ref{sec:conclusion} we draw conclusions about achieving an optimal degree of heterogeneity.

\section{Scientific and Technological Challenges}
\label{sec:challenges}

\begin{figure}[ht]
\centering
\includegraphics[width=.45\textwidth]{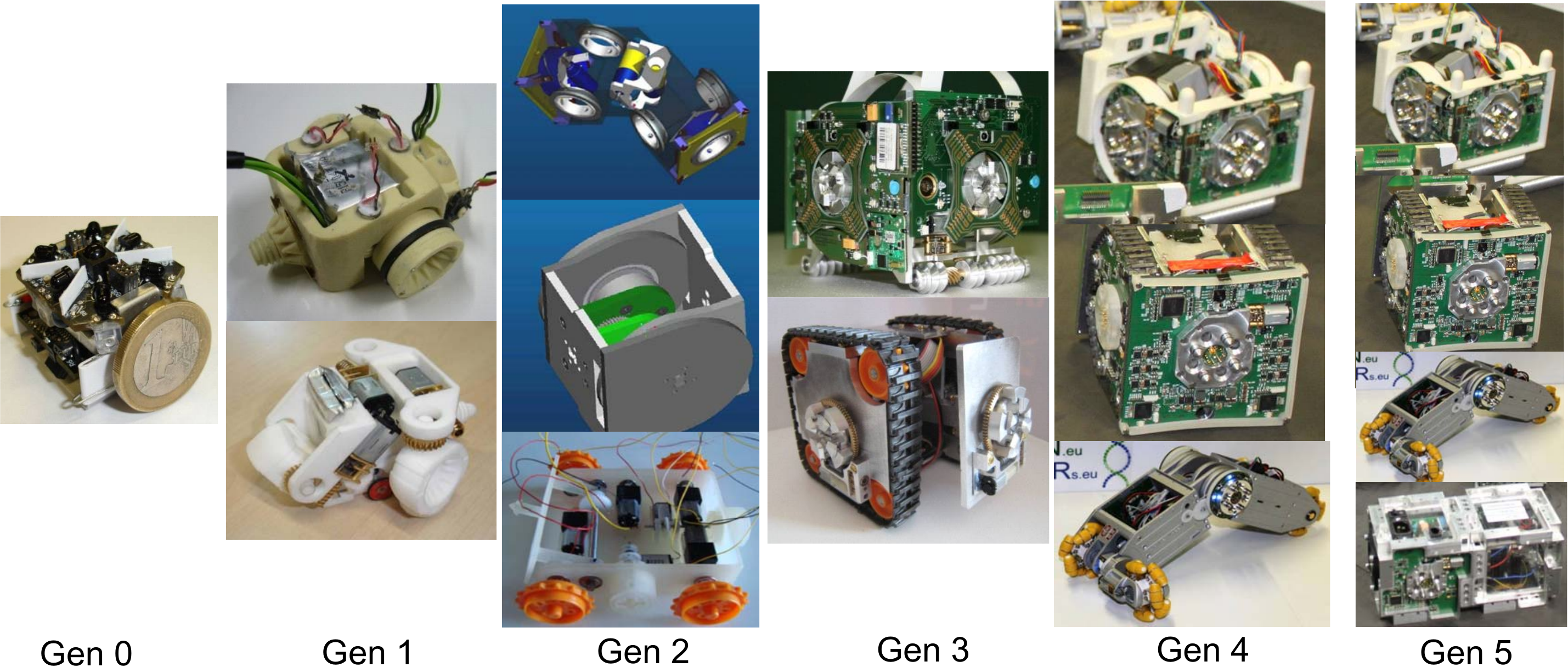}
\caption{\small From homogeneous to heterogeneous: five generations of platform development.}
\label{fig:6gen}
\end{figure}

The development started from the swarm concept and was originally intended for a small ($<\rm50~cm^3$) reconfigurable homogeneous swarm robot. From the original idea to the current stage, the platform development can be split into 6 generations, see Fig.~\ref{fig:6gen}. Original design (generation 0) started in 2007 from the Jasmine III robot, where the 1st generation extended this design for docking and 3D actuation capabilities. This is a homogeneous platform, following a chain-based reconfigurable design.

Since new partners with additional resources joined the project, the envisaged demonstration scenario increased its complexity. The first step into heterogeneity was made with the 2nd generation design, which is primarily motivated by increased functional requirements, see Table~\ref{tab:generalReq}. After initial topological studies~\cite{Kernbach08_2}, it became clear that 1DoF  (vertical and/or 180$^\circ$ rotational DoF) are enough for 3D actuation. However this system is not optimal for different required tasks. The proposal of making a complex universal homogeneous platform essentially increased the risk of particular or full failure due to complexity. Thus, the following factors: (a) different requirements of strong/weak DoFs in 2D and 3D mode, (b) geographic distribution of partners where a close collaboration became problematic, (c) increased risks of failure of a particular platform influenced the initial decision to split a design from one into two platforms.
\begin{table}[ht]
\centering
\begin{tabular}{p{2.0cm}@{\extracolsep{3mm}}
p{1.0cm}@{\extracolsep{3mm}}
p{1.0cm}@{\extracolsep{3mm}}
p{1.0cm}@{\extracolsep{3mm}}
p{1.0cm}@{\extracolsep{3mm}}
p{1.0cm}@{\extracolsep{3mm}}
}  \hline
Type & 1st Gen& 2nd Gen & 3rd Gen & 4th Gen & 5th Gen\\\hline
2D Locomotion & + & + & + & +& +\\
Autonomous Docking & + & + & + & +& +\\
3D DoF & 1 & 2 & 3-4 & 3-4 & 3-4\\
3D Movement & - & - & + & + & +\\
Power Bus & + & ++ & ++ & ++ &++\\
Power trophallaxis & - & - & + & + &+ \\
Commun. Bus & +  & + & ++ & ++ & ++\\
N of sensors & 10 & 22 & 22 & 22 & 22\\
Passive Modules  & - & - & - & - & +  \\
Risks & - & + & ++ & +++ & ++++  \\
Demo scenarios & 3D Aggre-gation & Artificial Organisms & NIST Benchmarks & Grand Challenges & Key Highlights\\
\hline
\end{tabular}
\caption {\small General functional requirements, first published in~\cite{CD11}. \label{tab:generalReq}}
\end{table}

These factors remained valid on all later stages, however we also encountered new bottlenecks. For example, during the third generation design, the conflict between the weight and available on-board power leaded to the decision to introduce the third platform (the active wheel), during the fourth generation -- the conflict between the number of modules and developmental plasticity leaded to the decision to introduce additional modules (passive modules). Thus, each new platform either resolves a constraint or optimizes a performance of previous solutions.

Each year, the funding agency and project reviewers increased the level of challenges for the final demonstration even more, from originally intended aggregation in 3D up to the "Key Highlights", see Table~\ref{tab:generalReq}. This essentially increased the impact of the projects, but also the complexity of platforms and risks of failures, see Table~\ref{tab:complexity}.
\begin{table}[ht]
\centering
\begin{tabular}{p{1.0cm}@{\extracolsep{3mm}}p{1.0cm}@{\extracolsep{3mm}} p{1.0cm}@{\extracolsep{3mm}}
p{2.0cm}@{\extracolsep{3mm}}
p{2.0cm}}
  \hline
Type & 1st Gen& 2nd Gen & 3rd Gen & 4th Gen \\\hline
MPU       & AVR   & ARM7  & ARM11+2 Cortex M3 & 2x-core Blackfin + 4 MSP\\
MIPS      & 12     & 36-80  & 1100                 &  3100\\
Memory  & 16kB & 128kB  &1mB                  & 64mB\\
Power    & 1,4W/h& 4W/h & 13W/h               & 33W/h \\
OS        & --       & --        & freeRTOS          & $\mu$Linux\\
Bus       & --       & CAN     & CAN/FlexRay     & Ethernet \\
\hline
\end{tabular}
\caption {\small Several characteristics of the platforms. \label{tab:complexity}}
\end{table}
To some extent, the complexity between each generation increased by the factor of 3--4 in such criteria like on-board energy, computation capabilities or software architecture. Scientific and technological challenges, addressed by each design generation (expressed in terms of demonstration scenarios), vary from the aggregation in 3D up to complex ecologies with different robot species.

\textbf{1. Aggregation:} Docking of modules into 3D structures and a collective actuation within such structures.

\textbf{2. Artificial Organisms:} 3D structures possess homeostatic regulation, evolutionary framework, distributed control system, aggregate/disaggregate into different multi-robot organisms.

\textbf{3. NIST Benchmarks:} This scenario is motivated by benchmarks, developed in NIST, USA. Experiments in different environments should test and evaluate such parameters of robot organisms like flexibility, collective actuation, degree of adaptability, cognitive capabilities~\cite{Kornienko_S05a}.

\begin{table}[ht]
\centering
\begin{tabular}{p{2.0cm}@{\extracolsep{3mm}}p{5.5cm}}
  \hline
Feature & Description \\\hline
1. Structural and Functional Diversity   &
Combination of different heterogeneous modules improves performance of common system and allows such a functionality, which was not possible for any of homogeneous elements.\\
2. Advanced Reliability & A high number of heterogeneous modules increase robustness and reliability of the system.\\

3. Adaptive (evolutionary) Structural and Functional Properties & High developmental plasticity enables short- and long-term adaptive and evolutionary processes. This results in a substantial increase of performance and efficiency in a predictably and unpredictably changing environment.\\
\hline
\end{tabular}
\caption {\small Key challenges addressed by heterogeneous platforms. \label{tab:challenges}}
\end{table}

\textbf{4. Grand Challenges:} These are two complex high-impact scenarios. In the first scenario 100 robots should survive for 100 days without any human assistance in a changing environment. The second scenario was intended to explore evolutionary conditions similar to those that leaded to the appearance of multi-cellular species during the Cambrian explosion.

\textbf{5. Key Highlights:} These demonstration scenarios represent further refinement of the Grand Challenges with three or more dedicated experiments highlighting key scientific and technological challenges of artificial organisms.

Thus, each time we observe an increasing pressure on the number and complexity of different experiments. Generalizing the main challenges from all these scenarios, the advantages of heterogeneous platforms are located around three following aspects: \emph{diversity}, \emph{reliability} and \emph{adaptivity}, see Table~\ref{tab:challenges}. The following sections introduce these heterogeneous solutions and demonstrate their specialization and common elements.

\section{Technological Solutions for Large-Scale Heterogeneous Self-Reconfigurable Robots}
\label{sec:solutions}

Four platforms have been developed within the scope of this discussion, see Table~\ref{tab:overview}, which are specialized in different DoF in 2D and 3D mode as well as optimized for different tasks.
\begin{table}[htbp]
\centering
\begin{tabular}{
p{1.6cm}@{\extracolsep{3mm}}
p{1.2cm}@{\extracolsep{4mm}}
p{1.2cm}@{\extracolsep{3mm}}
p{1.2cm}@{\extracolsep{3mm}}
p{1.3cm}}
\hline\noalign{\smallskip}
\multicolumn{5}{c}{\textbf{Common Elements}}\\\hline
Docking          & \multicolumn{4}{p{5cm}}{genderless, $90^\circ$ symmetric, Ethernet and power buses}\\
Energy Bus       & \multicolumn{4}{p{5cm}}{8~A at 25~V, current limiter at each module, controllable power distribution}\\
Battery          & \multicolumn{4}{p{5cm}}{6 LiPo cells, 1,400~mAh, total nominal voltage: 22,2~V}\\			
Communication    & \multicolumn{4}{l}{Internal Ethernet bus, ZigBee, local IR, sound} \\
Electronics      & \multicolumn{4}{p{5cm}}{Universal electronic architecture, the same set of sensors} \\
Software         & \multicolumn{4}{l}{Universal software framework}\\[2mm]\hline
\multicolumn{5}{c}{\textbf{Heterogeneous Elements}}\\[1mm]
                         & Scout Platform         & Backbone Platform         & Active Wheel           & Passive Modules \\\hline
Specialization      & exploration tasks & strong 3D actuation & transpor-tation tasks & e.g. add. power, electronics \\
Locomotion		     & tracked locomotion  & nearly Omni-directional   & Omni-directional       & none\\
$N$ of Docking Elements & 4 & 4 & 2 & 1 - ...\\
Speed, loc.  		 & 12.5 cm/s 			    & 6 cm/s 		                  & 31 cm/s                  & none\\
DOFs of actuation & bending: $\pm90^\circ$ rot.: $\pm180^\circ$ & $bend./rot.:\pm90^\circ$  & bend./rot.: $\pm180^\circ$ & none \\
Torque 				 & up~to~4~Nm					    & up~to~7~Nm 		              & up~to~5~Nm               & none \\
Speed, act. 	     & $37.2^\circ$/s        & $90^\circ$/s 	              & $50^\circ$/s            & none \\
Volume               &     1356~cm$^3$      & 1157~cm$^3$                                   &  1470~cm$^3$           & \\
Weight               &     1000~g                & 1000~g                                   &  1550~g                     & \\
\noalign{\smallskip}\hline\noalign{\smallskip}
\end{tabular}
\caption{\small Overview of common (unified) and heterogeneous elements in the design of heterogeneous reconfigurable robots. \label{tab:overview}}
\end{table}
To make a joint operation of all these platforms possible, a number of common elements have been developed. The following sections provide a detailed sight into these elements.

\subsection{Common Elements: Unified Docking Mechanism}
\label{sec:dockingElement}

\begin{figure}[h!]
	\centering
		\includegraphics[scale=.3]{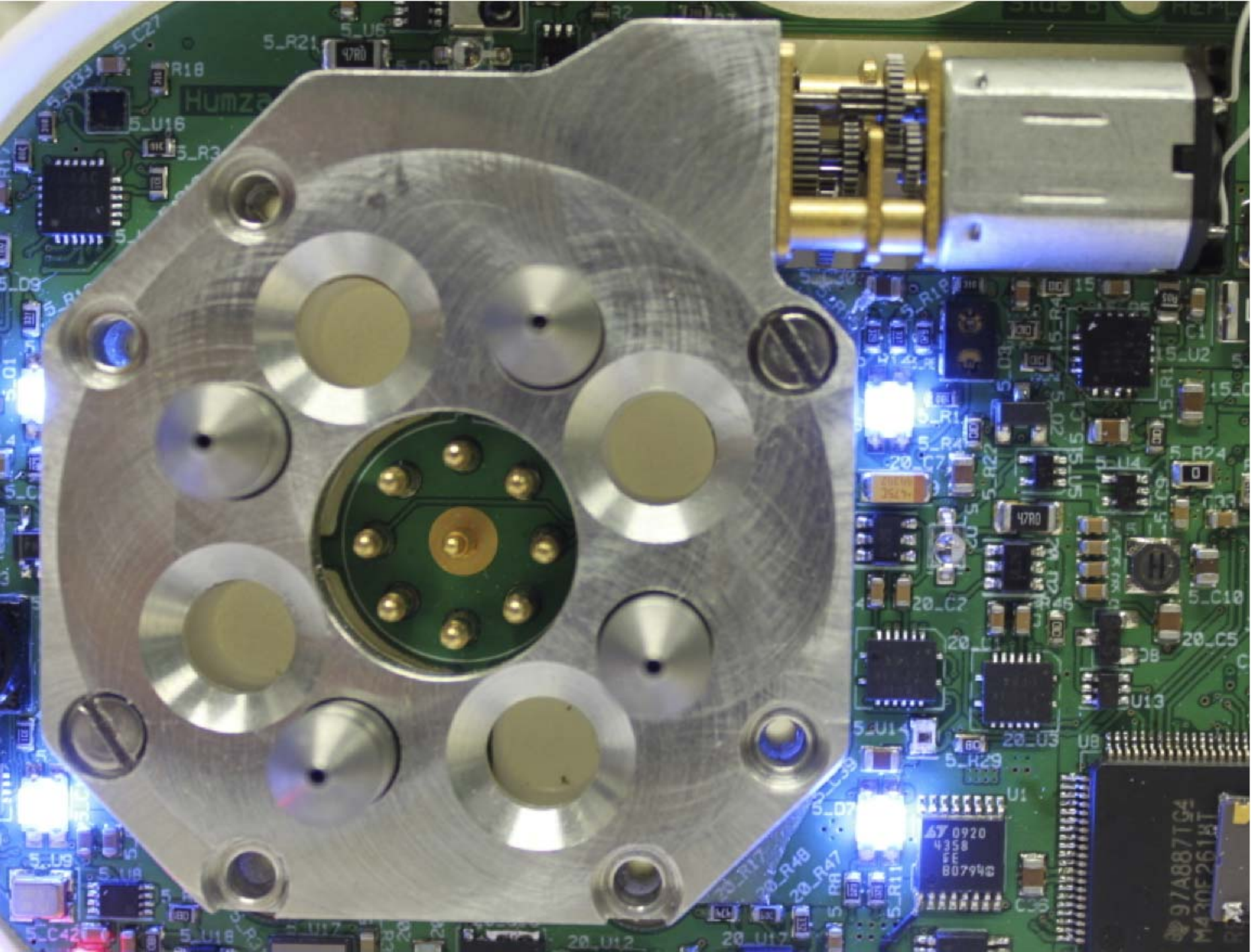}
		\caption{\small Unified homogeneous docking unit.\label{fig:dockingUnit}}
\end{figure}

A key element of heterogeneous robot platforms is the ability to dock mechanically and electrically to merge into a larger artificial multi-cellular organism. The docking design used in our modular robots is called CoBoLD (for Cone Bolt Locking Device, see Fig.~\ref{fig:dockingUnit}) and among other features it is genderless, $90^\circ$ symmetric and handles misalignment.

The design itself is based on 4 cone shaped, spring loaded bolts. During docking the bolts will be guided by chamber-bevels of the facing docking units inserted holes and finally locked inside by a hook. The locking mechanism uses a DC motor driven worm gear providing enough force to securely connect two robots. Due to the self-locking feature of a worm gear, no energy is required to hold the connection.

It is possible to push the bolts inside the docking unit resulting in a single boundary box regardless of the orientation or state of the locking mechanism. This can be important, if e.g. one robot is turned lying on its docking unit and another robot needs to dock. In the center of the docking unit, spring loaded electrical contacts are placed to assure an automatic connection of power and wired communication when two robots dock. The contacts are placed on the same PCB holding all the other electrical components on this robot side. This eases the electronic design since no cables or additional contact PCBs are required. The arrangement of these contacts plus a special switching circuit ensure that robots can dock no matter how they are oriented with respect to each other.

Several versions of CoBoLD are available, depending on the required functionality. For example, if active docking (the ability to lock bolts) is not required, a passive docking unit can be used, reducing the size of the assembly while still providing for all other mechanical and electrical features.

\subsection{Common Elements: Unified Energy Sharing System}
\label{sec:energy}

A ``unified energy sharing" system of a reconfigurable multi-robotic organism takes its inspiration from the ant's social stomach. It not only allows them to exchange power but also distributively and collectively monitors and regulates the dynamic power flow between the multiple power sources that are now connected in parallel through an organism's power sharing bus. A ``unified solution" is in-fact required to reduce the complexity of such a dynamically evolving system~\cite{Kernbach08online}. In principle, it provides a base platform to the application layer control routines to develop and evolve complex behaviors with the changing environment~\cite{Kornienko_S06b}, e.g., establishing energy homeostasis, power aware fault tolerance, dynamic power sharing, emergent phenomena based on energy distribution, etc~\cite{Kornienko_S04a}.

\begin{figure}[h!]
	\centering
		\includegraphics[scale=0.3, angle = 0]{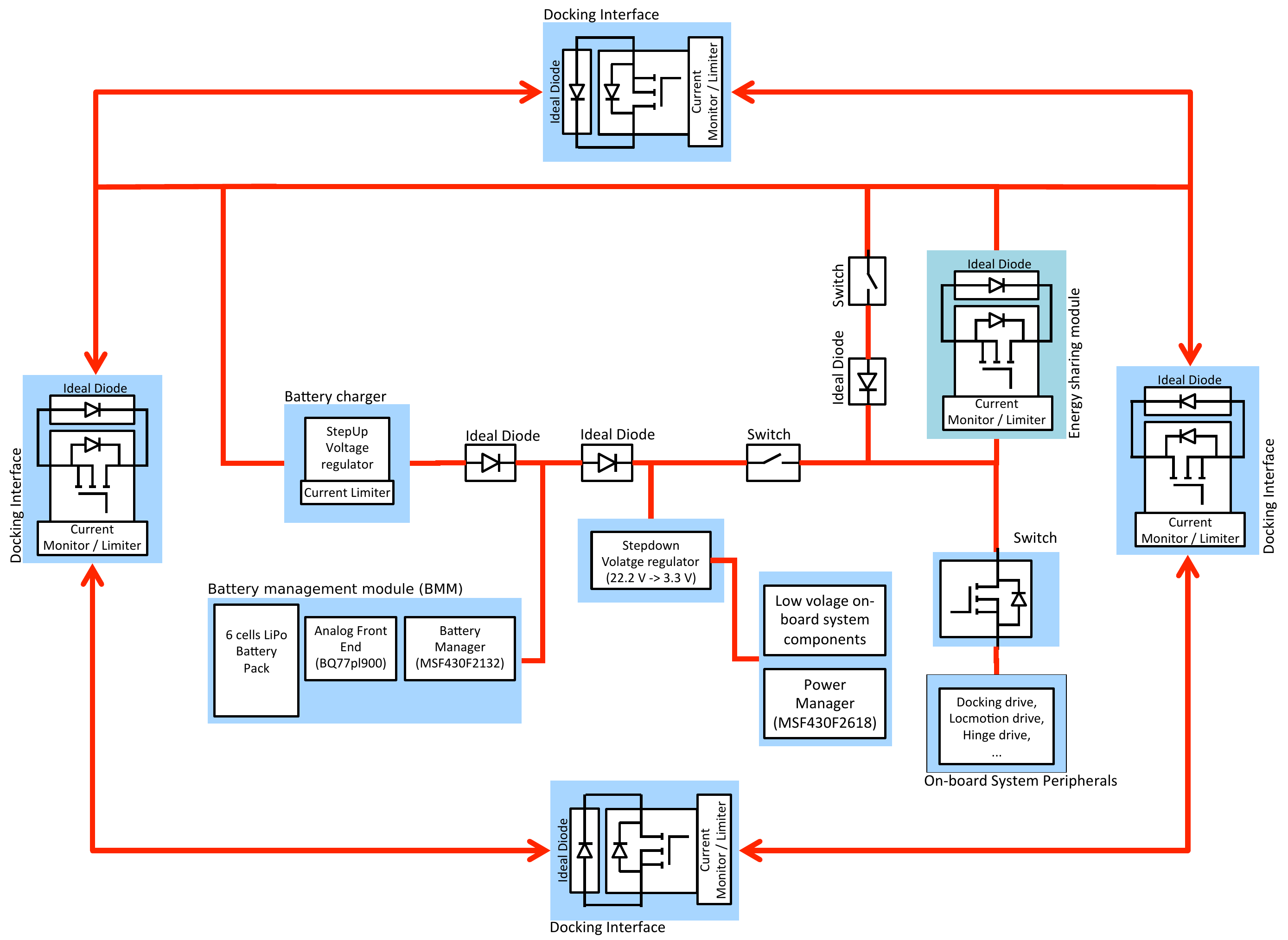}
	\caption{Block diagram of the power management system.	 \label{fig:PMblockDiagram}}
\end{figure}

Fig.~\ref{fig:PMblockDiagram} shows the block diagram of the proposed power management system of a single robotic module. The hardware design of the power management system includes a battery management module, a battery re-charging unit, ideal diodes for uni-directional low loss current flow, a system power manager, an energy sharing module, a high power control switch for powering on-board peripherals, and four docking interfaces one on each lateral side of the robot.

The unified energy system mainly consists of 4~docking interfaces and an energy sharing module. The on-board energy sharing module in its ON state allows an individual to share its battery energy with other modules. While in its OFF state, allows an inward current flow, provided that the on-board source voltage is lower than the voltage on the power bus. The docking interface on each side of the robot provides the on-board power manager with a means to control the outward flow of energy. Additionally, the current monitor at each docking interface and in the energy sharing module allows the system power manager to measure the current flow in either direction. This way, each individual distributively but collectively controls the current flow through the organism's power bus. Besides the common power management architecture, all platforms are equipped with the same custom designed battery packs.

\subsection{Common Elements: Electronic Framework}
\label{sec:electronics}

\begin{figure}[h!]
\centering
	\includegraphics[scale=.3]{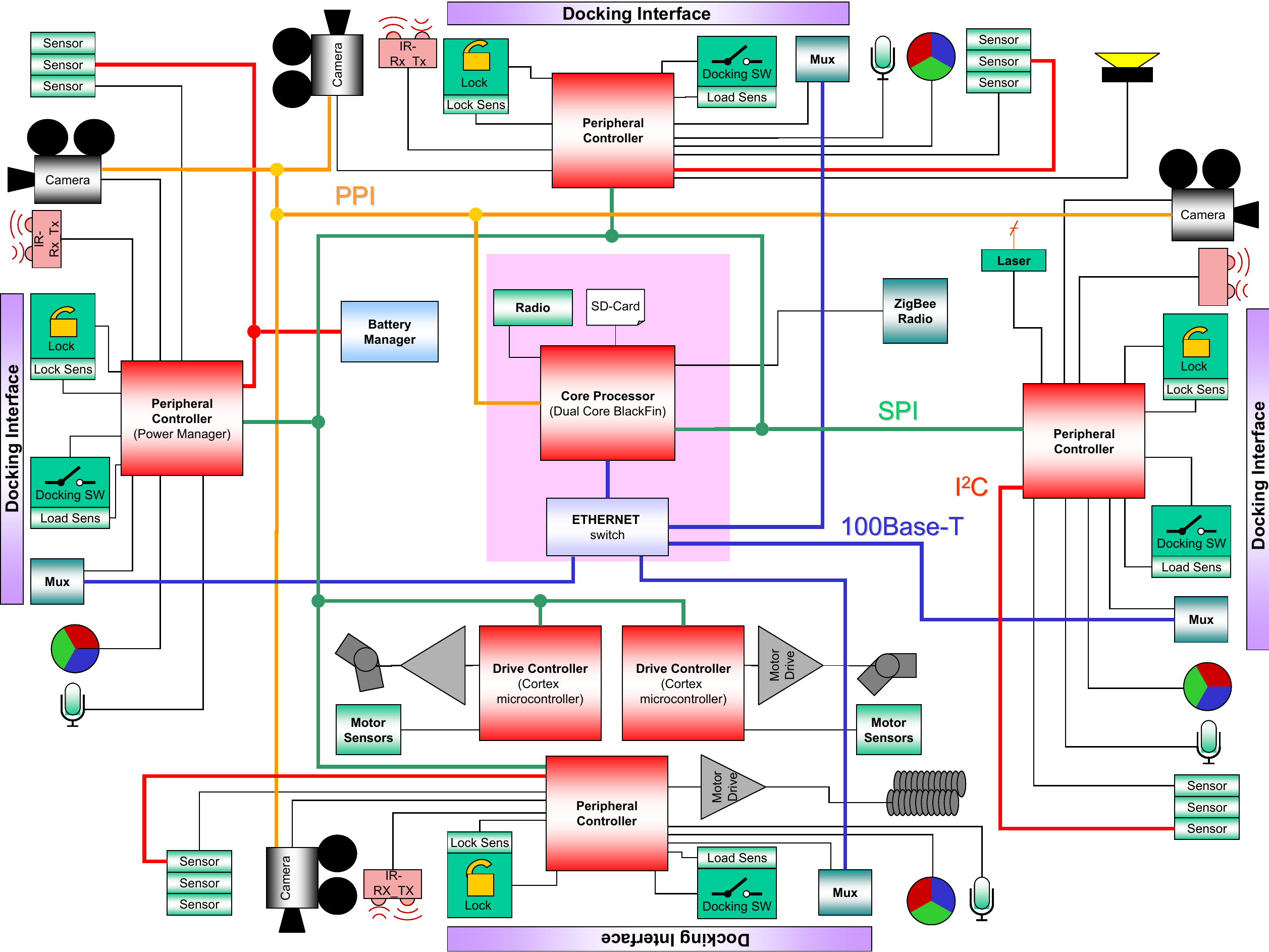}
\caption{\small Block diagram of the electronic architecture. }
\label{fig:Earchi}
\end{figure}

The general electronic architecture is shown in Fig.~\ref{fig:Earchi}.
The main processing unit of each of the robot platforms is a \textit{CM-BF561} drop-in module from \textsc{Bluetechnix} equipped with a dual core \textit{Blackfin} \textit{BF561} $\mu$controller from \textsc{Analog Devices}. Via a Serial Peripheral Interface (SPI), this unit is attached to 4~peripheral $\mu$controllers (\textit{MSP430F2618} from \textsc{Texas Instruments}), which are responsible for sensor data acquisition, low level actuator control and processing in order to take off burden from the main processing unit. These peripheral controllers  mainly serve as an interface between the application software routines running on the \textit{Blackfin} processor and the system components. All $\mu$controllers and corresponding peripheral elements are placed on separate PCBs, installed on each side of a robot module. Each of these PCBs have in common certain sensors and actuators (e.g. docking sensors and actuators, microphones, RGB-LEDs, etc.), but may in addition take over specalised tasks (e.g. 2D or 3D locomotion, ZigBee radio interface, etc.). For 3D actuation, up to two additional \textit{LM3S8962} \textit{Cortex} $\mu$controllers (\textsc{Luminary Micro}) are integrated to permit high-performance brushless DC-motor control. On each of the PCBs, a local I$^2$C bus is implemented for interfacing the local controller with the respective sensors. In addition, a global I$^2$C bus has been implemented to facilitate multi-master communications between the peripheral controllers.

\subsection{Heterogeneous Elements: Backbone Platform}

\begin{figure}[h!]
\centering
	\includegraphics[scale=.25]{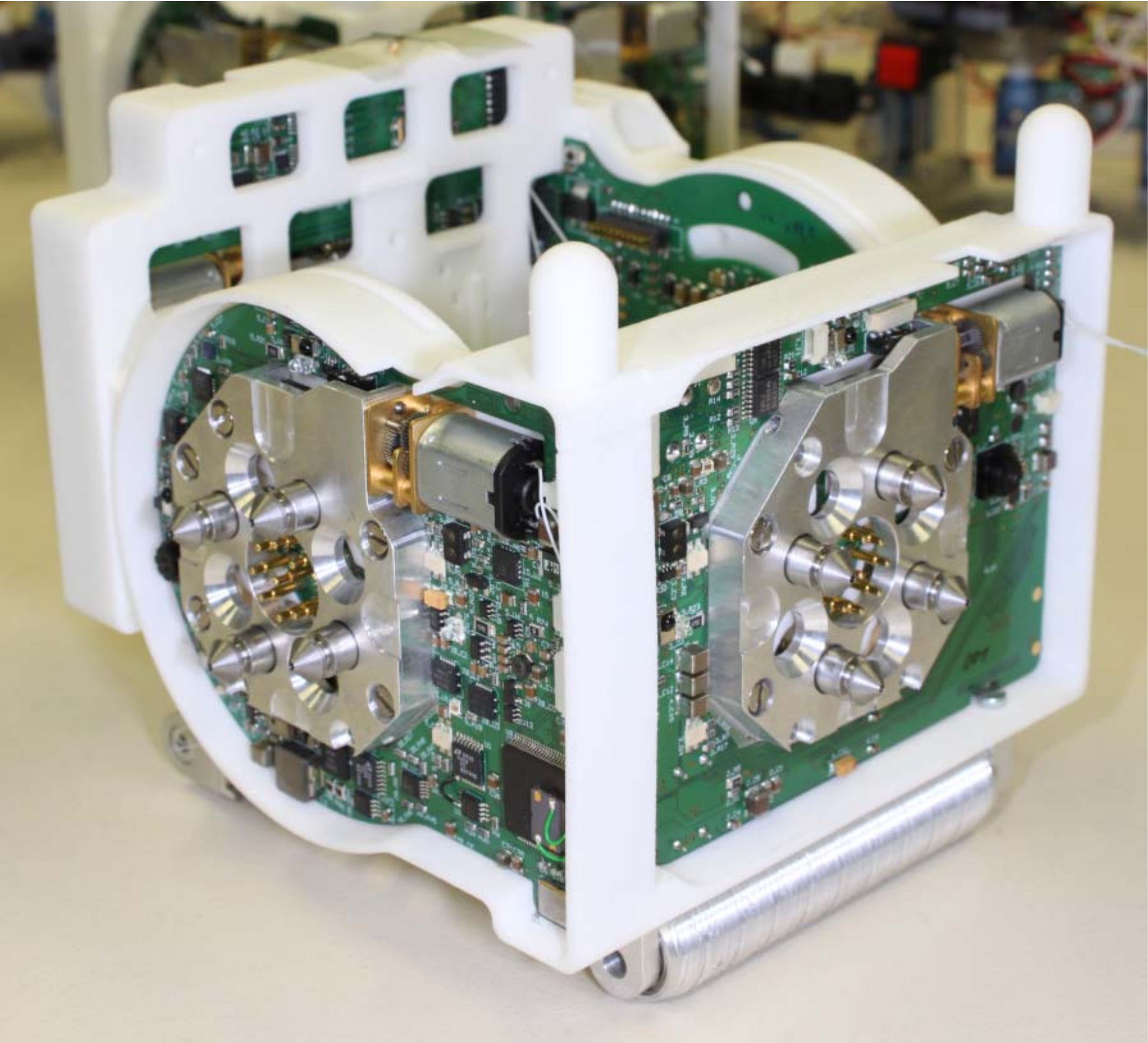}
\caption{\small A rapid prototype version of the Backbone platform. \label{fig:dummyRobot}}
\end{figure}

The backbone platform, see Fig.~\ref{fig:dummyRobot}, is specialized in strong 3D locomotion and actuation capabilities in the organism mode. The robot possesses a strong brushless drive capable of lifting several robots. A single DOF approach reduces the cost and since the main motor is a rotation-symmetric subassembly, the available space inside is better suited to support one strong motor instead of several smaller ones. To compensate the reduced number of DOFs, the frame of the robot uses two L-shaped halves, which can be rotated against each other. This allows for lifting or rotating connected robots depending on their docked position. A symmetric, genderless, active docking was chosen to eliminate the need to search for compatible docking interfaces. This increases the diversity of possible organism structures that can be built. In order to use the feature of the four mounted docking units, which not only provide for a stable connection but also for power and communication transfer, a special 2D drive was included allowing the robot to additionally move sideways. The robot can move freely as a single unit and it is possible to use the 2D drives of several connected robots to drive the organism. A cubical shape was chosen because in our opinion it combines best the requirements for a small sized swarm robot and the requirements for modular self-reconfigurable robots, where a symmetric shape would simplify the reorganisation of an organism.

So far, we demonstrated that the 2D drive can move a single robot in all directions, all docking units can be used to connect the robots. The cubical shape allows for docking no matter how the module is oriented. In addition, docking between the different platforms is possible. General functionality of the 3D drive was demonstrated in the 3rd generation module, the implementation into the current design of the robot is ongoing work. Due to the internal design, bending and rotating is limited to $180^\circ$.

\subsection{Heterogeneous Elements: Active-Wheel Platform}
\label{sec:activeWheel}

\begin{figure}[h!]
\centering
\includegraphics[scale=.3]{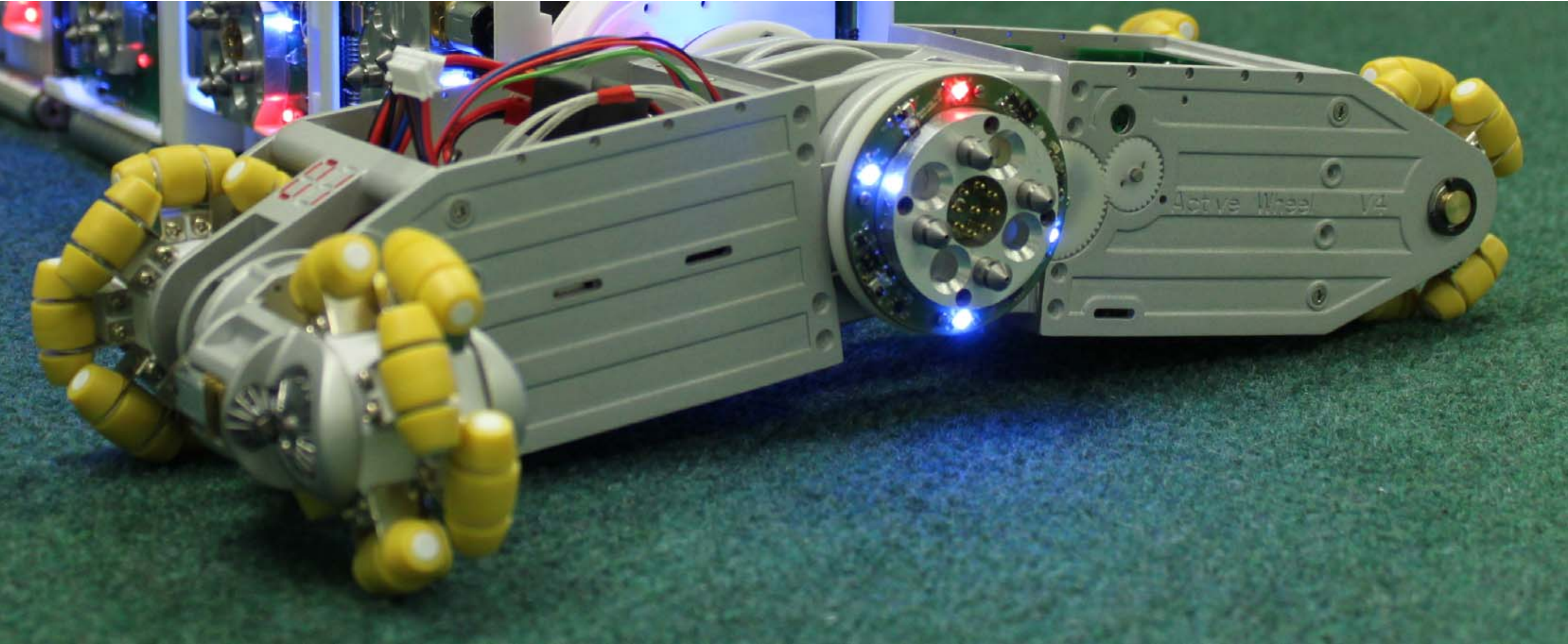}
\caption{\small Active wheel. 	\label{fig:activeWheel}}
\end{figure}

The Active Wheel, see Fig.~\ref{fig:activeWheel}, is a fully functional platform, specialized in transportation tasks. It was designed to carry and transport an organism consisting of several Scout or Backbone robots in a most energy-efficient way. The robot is able to approach the requested module/organism and dock to it. The current design consists of two symmetrically arranged arms. On each end two $90^\circ$ shifted Omni-Wheels are attached, which together form the locomotion mechanism of the robot. These arms are connected via a $180^\circ$ turning hinge. Docking to other robots is provided by 2 docking elements placed on the same axis as the hinge. To allow docking in any position or height, the docking elements can rotate, actuated by two separate motors. For the production of the prototypes a vacuum-casting process was chosen. This method allows to produce multiple chassis in one working step simultaneously. Thereby, the manufacturing cost of the mechanics was reduced by 40~\% compared to standard rapid prototyping.

The chassis of the robot has two cavities, which are ideal to place the main electronics. On the one hand, this configuration protects the sensitive components from damage; on the other hand, it allows placing all sensors outside in order to perceive the environment. However, this sepaarates the electronics into different areas and increases the number of necessary printed circuit boards. In total, 15 boards were installed on each robot, whereby 6 of them are used as a sensor for detecting the motor speed for driving and docking.

\subsection{Heterogeneous Elements: Scout Platform}
\label{sec:ScoutRobot}

\begin{figure}[h!]
	\centering
	\subfloat[]{\includegraphics[scale=.4]{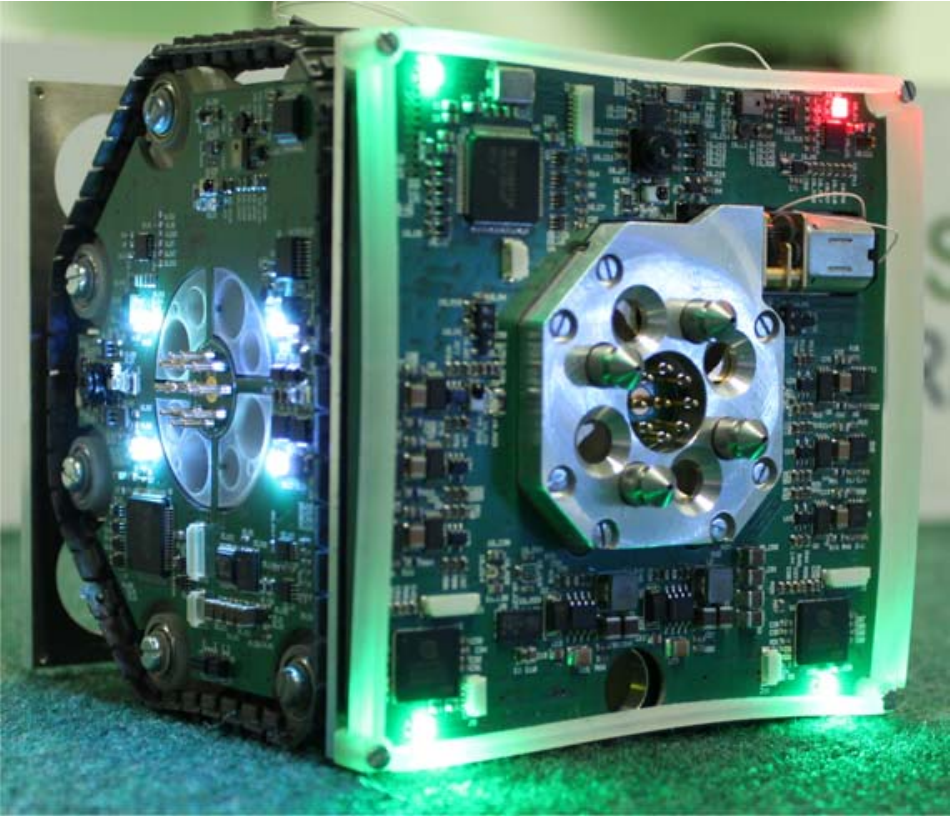}   
\label{fig:sssa-01} }		
	\subfloat[]{\includegraphics[scale=.22]{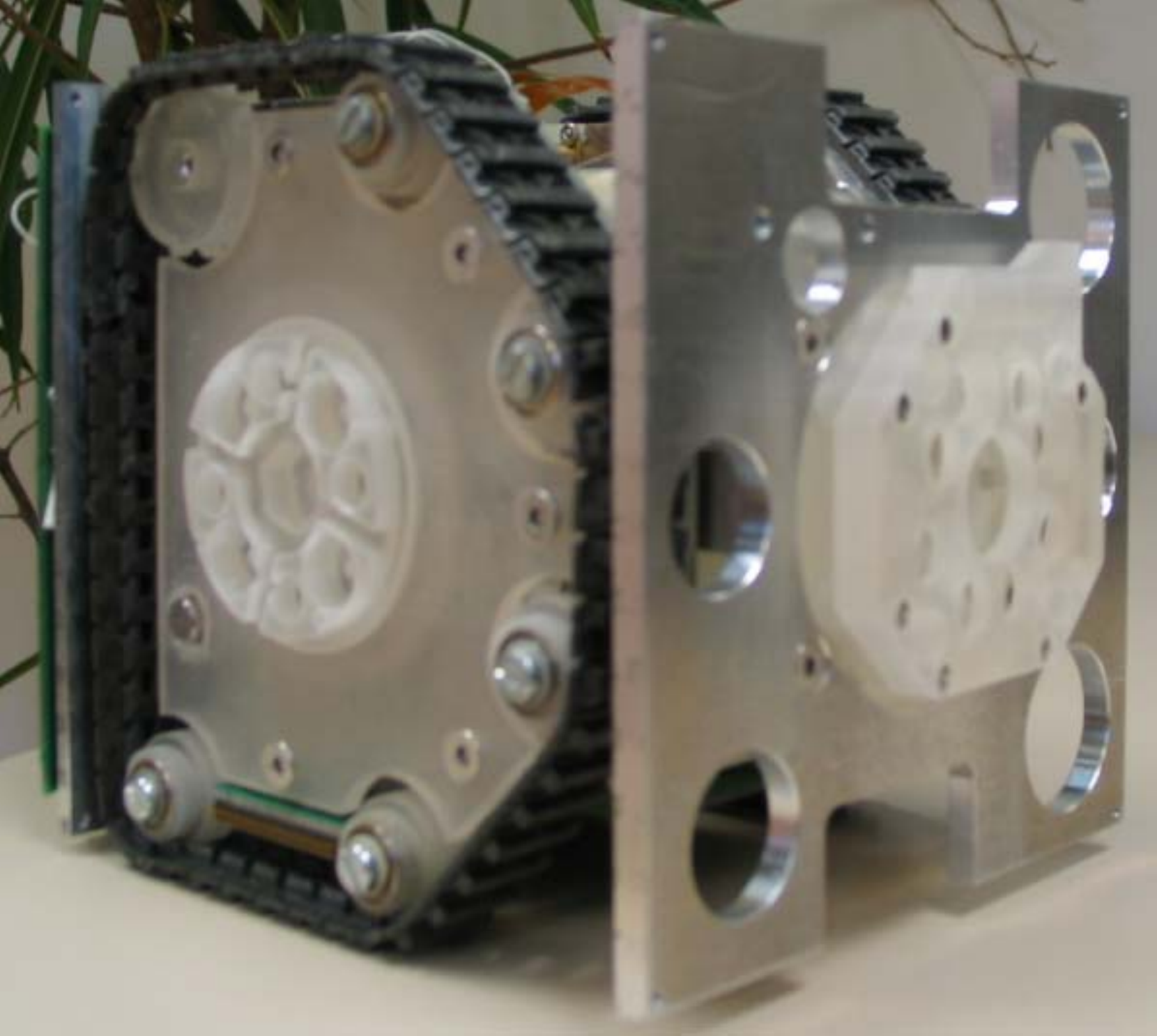}   
\label{fig:sssa-02} }		
	\caption{\small The Scout robot platform: (a) front and right sides of the Scout robot, (b) rear and left sides of the Scout robot with stripped off PCBs.}
	\label{fig:sssa}
\end{figure}

The Scout robot, see Fig.~\ref{fig:sssa}, is specialized in fast locomotion on challenging terrains for exploration tasks. Therefore, the platform is equipped with additional sensors compared to the other platforms in order to scan the surroundings and the floor. In particular, the Scout robot has two laser-camera systems, on its front and rear to provide far and short range obstacle detection. The Scout robot's locomotion is based on tracks. Fast locomotion for the exploration on rough terrains is more important for the Scout robot than slow and precise locomotion for aggregation and docking alignment, compared to the Backbone Robot. Tracks enable the Scout robot to move forward, backward, turn left, turn right and turn on its axis. Moreover, the Scout robot can move on rough terrains, climb slopes and overcome small obstacles. The speed of locomotion is more than one body length per second. Thanks to continuous elastic tracks, the robot is able to perform locomotion even after overturning accidentally.

Similar to the Backbone robot, the Scout robot has four docking units, one on each lateral side wall. The docking units are centered on the walls of the robot so that modules can dock regardless of their orientation. This platform has 2 DOF, bending and rotation, with a maximum torque of 4,7~Nm. The bending DOF ($\pm90^\circ$) allows the robot's rear wall to lift up and down while the rotational DOF ($\pm180^\circ$) allows the docking unit on the robot's left side to rotate along its axis. The 3D DOF has lower torque than the Backbone robot since the main role of the platform is exploration and not macro-locomotion. In any case, the Scout robot is able to lift two other modules. The Scout robot, with higher DOF and lower torque compared to the Backbone robot, performs best if docked to the tail or to the head of an organism to scan the environment.

\subsection{Heterogeneous Elements: Passive Modules}
\label{sec:structuralElements}

Passive modules will increase the heterogeneity degree of an organism. Depending on the design, a passive module can have the same shape like a robot to match inside an organism. It is equipped with at least one docking unit robots can connect to and e.g. is equipped with additional power packs or electronics. Such a passive module could increase for example the operation time of an organism or add additional features like specialized sensors and high-resolution cameras.

Since passive modules can be used in place of a robot, each version of must be able to interact with a docked robot. Passive modules are therefore equipped with the same communication interfaces (wired, wireless) and power circuits like a robot to assure communication is forwarded and power can be distributed. Furthermore, since passive modules interact with the organism (e.g. processing camera pictures in a module with a specialized FPGA mounted) a connection to the internal communication bus is available.

\subsection{Software Platform}
\label{sec:software}

\begin{figure}[h!]
	\centering
		\includegraphics[width=0.475\textwidth]{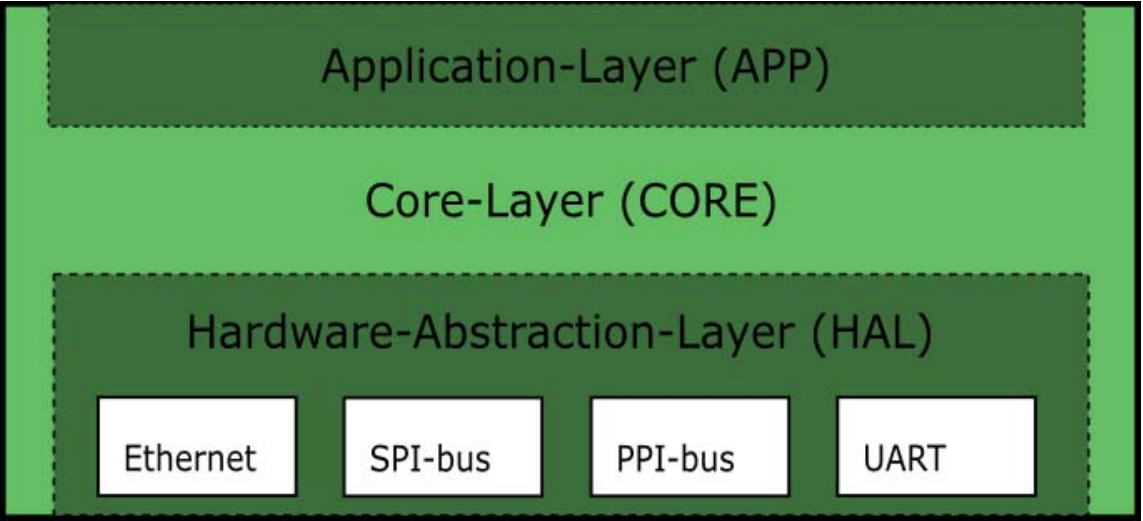}
		\caption{\small Layered architecture of the software framework.}
	\label{fig:swarchlayers}
\end{figure}

Although we have heterogenous platforms, unified interfaces are necessary for the ease of use of the resulting software. It needs to be stated here that the choice to go for a Blackfin DSP as main CPU and to have several minor processing units already implies some constraints for the general software architecture. It is intended to have a layered architecture consisting of three layers. Those three layers as well as their composition with respect to each other can be seen in Fig.~\ref{fig:swarchlayers}.

Due to the way the $\mu$Clinux OS running on the Blackfin DSP deals with scheduling threads on both cores it is necessary to have several heavy weight processes within the architecture. Therefore the need for inter-process communication arises, which will be handled within the core layer (CORE). This layer will mainly be comprised of a Daemon which monitors the communication among all the processes. The application layer (APP) is to hold controllers and high-level behaviors. It is able to utilize the functionality of the layers residing under it. The hardware-abstraction layer (HAL) underneath the core layer will deal with all the physical interfaces the Blackfin provides. It is meant to provide further hardware abstraction and to hide the complexity of the underlying hardware up to a certain extent. Each one of the white boxes in Fig.~\ref{fig:swarchlayers} denotes a heavy weight process and encapsulates all the functionality of the according hardware interface. The two most complex parts here will be the Ethernet interface and the SPI-bus interface.

\begin{figure}[h!]
	\centering
		\includegraphics[width=0.475\textwidth]{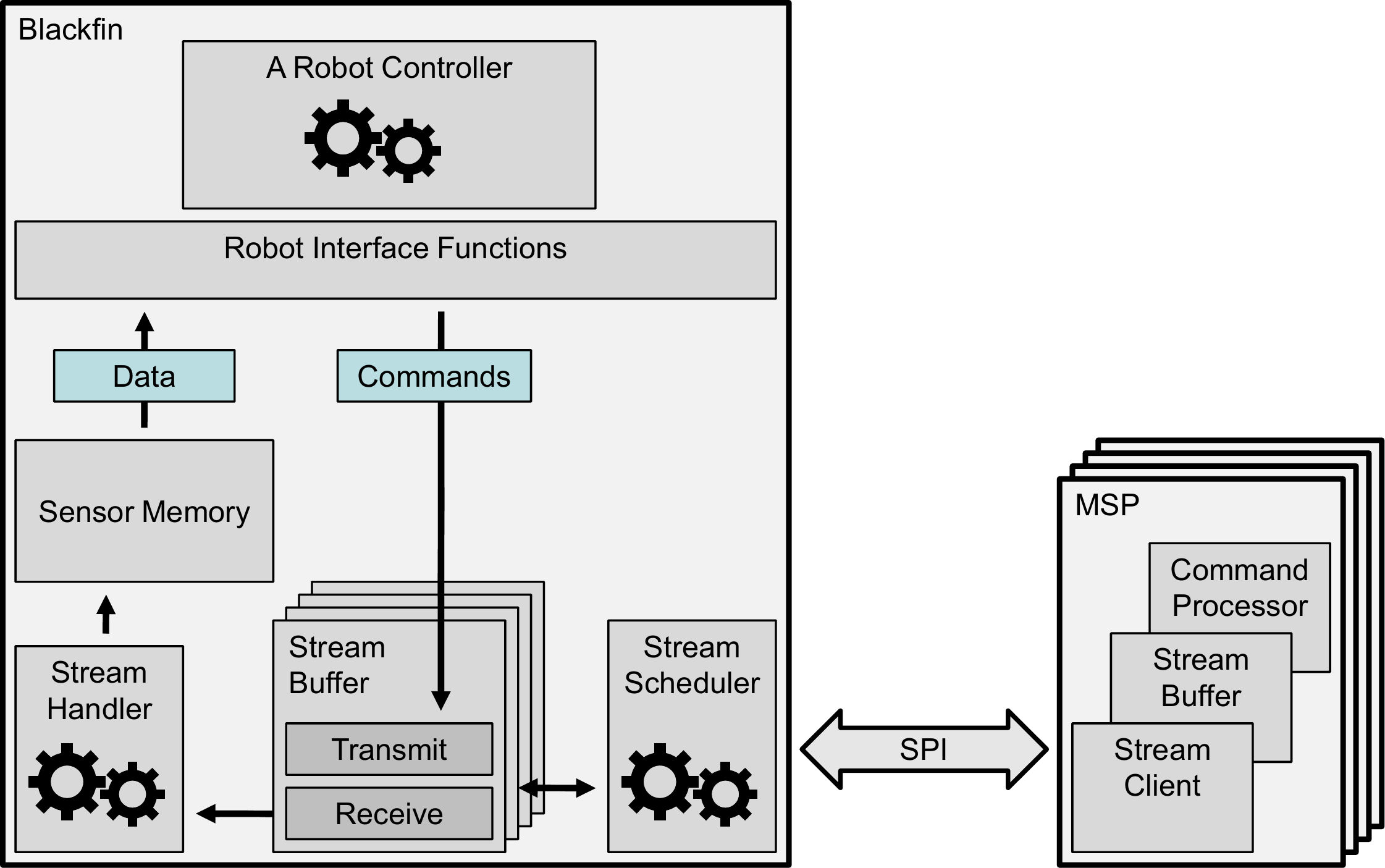}
		\caption{\small Architecture of the SPI-software.}
	\label{fig:coresw}
\end{figure}

The SPI-bus deals with most of the local functionality of the robots, as it connects the MSP controllers to the Blackfin. It therefore needs to be explained here in detail. Several tasks have to be performed within the SPI-process in Fig.~\ref{fig:swarchlayers}: Initiating communication to the MSP controllers, scheduling the use of the bus to ensure proper bandwidth allocation and error handling as well as correction. Furthermore the differences within the electronics among the three different platforms have to be taken into account here. The software framework therefore supports all three types of robots and allows a reusable design of controllers for the robots. For all robot types a common interface exists, which allows to write controllers for all types of robots. The interface can distinguish between the types and is extendable to new functions or future robot design. One part of the SPI software is run on the MSP controllers. The other part resides on the Blackfin within $\mu$Clinux. Fig.~\ref{fig:coresw} shows the principle of the SPI-software and the intended data flow on the robots.

The SPI communication is a bidirectional stream and is driven by the Blackfin. Additionally, the MSP can trigger a read, if new data is available. The received sensor data on the Blackfin is pushed into a sensor memory, where controllers can access the data via the interface later on. Vice versa, interface commands sent by a robot controller are transmitted via SPI to the designated sub processor.

Due to the different robot type based characteristics an adjusted hardware abstraction layer is implemented for each robot. This layer is specialized for each robot type and is abstracted to the user for better use. A common interface unifies the functionality of all three robot platforms into a collection of higher level functions. Most of the functions can be used on all three robots (e.g. sensor access, communication). Some of the functions, like locomotion, are specialized to the robots, but can be abstracted to simple common locomotion directives, like MOVE or TURN, which are valid for all types again. Via the interface, a controller can distinguish on which robot platform it runs and call the according function (e.g. four-wheeled locomotion for the Active Wheel).

\section{Manufacturing Issues}
\label{sec:manufacturing}

\begin{figure}[h!]
	\centering
		\includegraphics[width=0.45\textwidth]{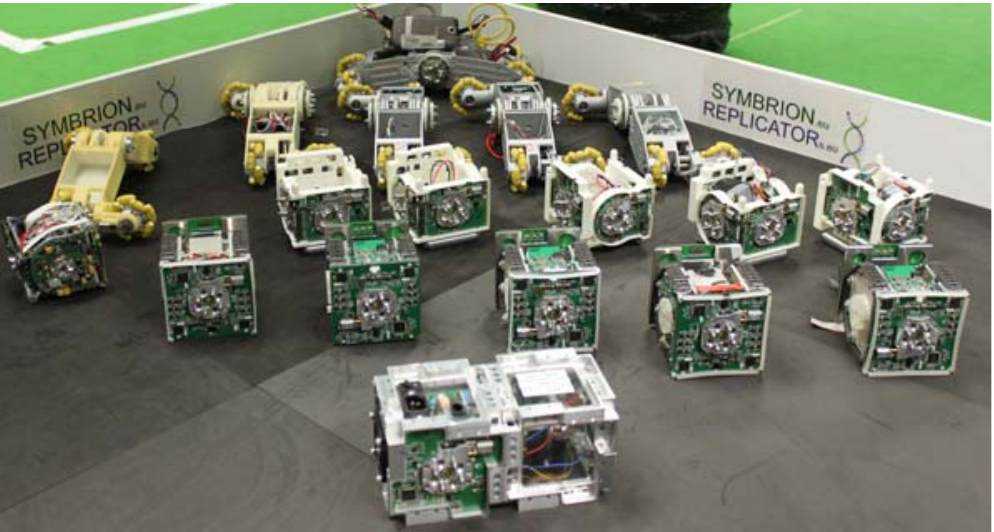}
		\caption{\small Manufactured heterogeneous robot platforms. \label{fig:15robots}}
\end{figure}

Manufacturing of a large number of modules is a critical process. Currently we produced 15 prototypes of fully functional robots (5 pieces of each platform) by using RP technology, see Fig.~\ref{fig:15robots}. This intermediate run is used for improvements, resolving hardware problems as well as for developing the software framework.
In the final run, it is planned to produce a large scale swarm (100+ robots). Manufacturing and maintaining such a number of complex modules involves new aspects into design of platforms: simplicity of assembly, protection of electronics, allocation of funding for maintaining these robots and for performing experiments. Generally, PCBs, SMD assembly and mechanical elements can be ordered. However, a final assembling and testing of modules represents a highly challenging problem for academic partners.

\section{Performed Experiments}
\label{sec:experiments}

The following two scenarios aim at a demonstration of simplest qualitative cases leading to improved performance and increased reliability of a heterogeneous system.

\subsection{Improving Performance of a Heterogeneous Organism}

\begin{figure}[h!]
	\centering
	\subfloat[]{\includegraphics[width=0.24\textwidth]{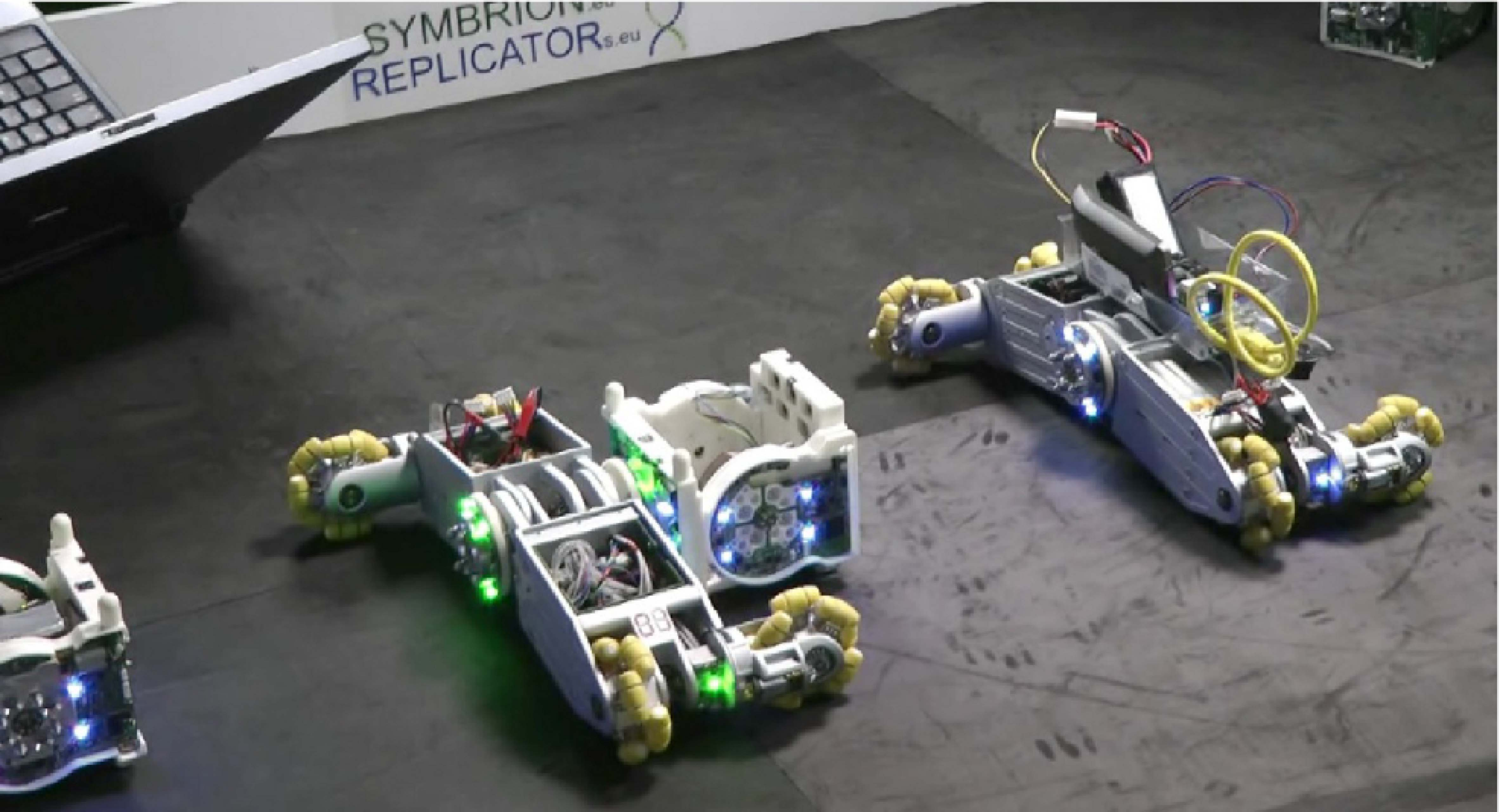}}
	\subfloat[]{\includegraphics[width=0.24\textwidth]{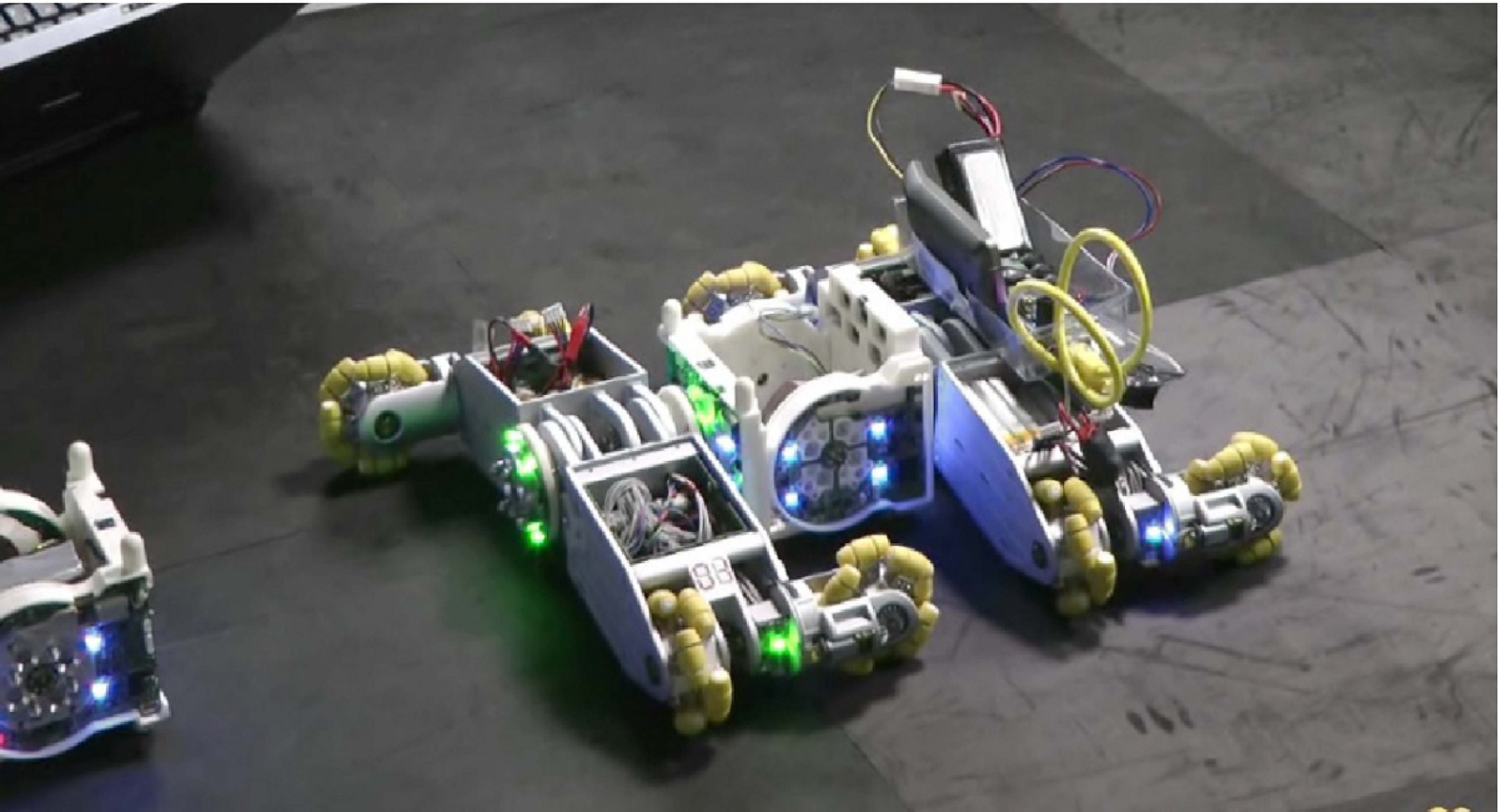}}
	\newline
	\subfloat[]{\includegraphics[width=0.24\textwidth]{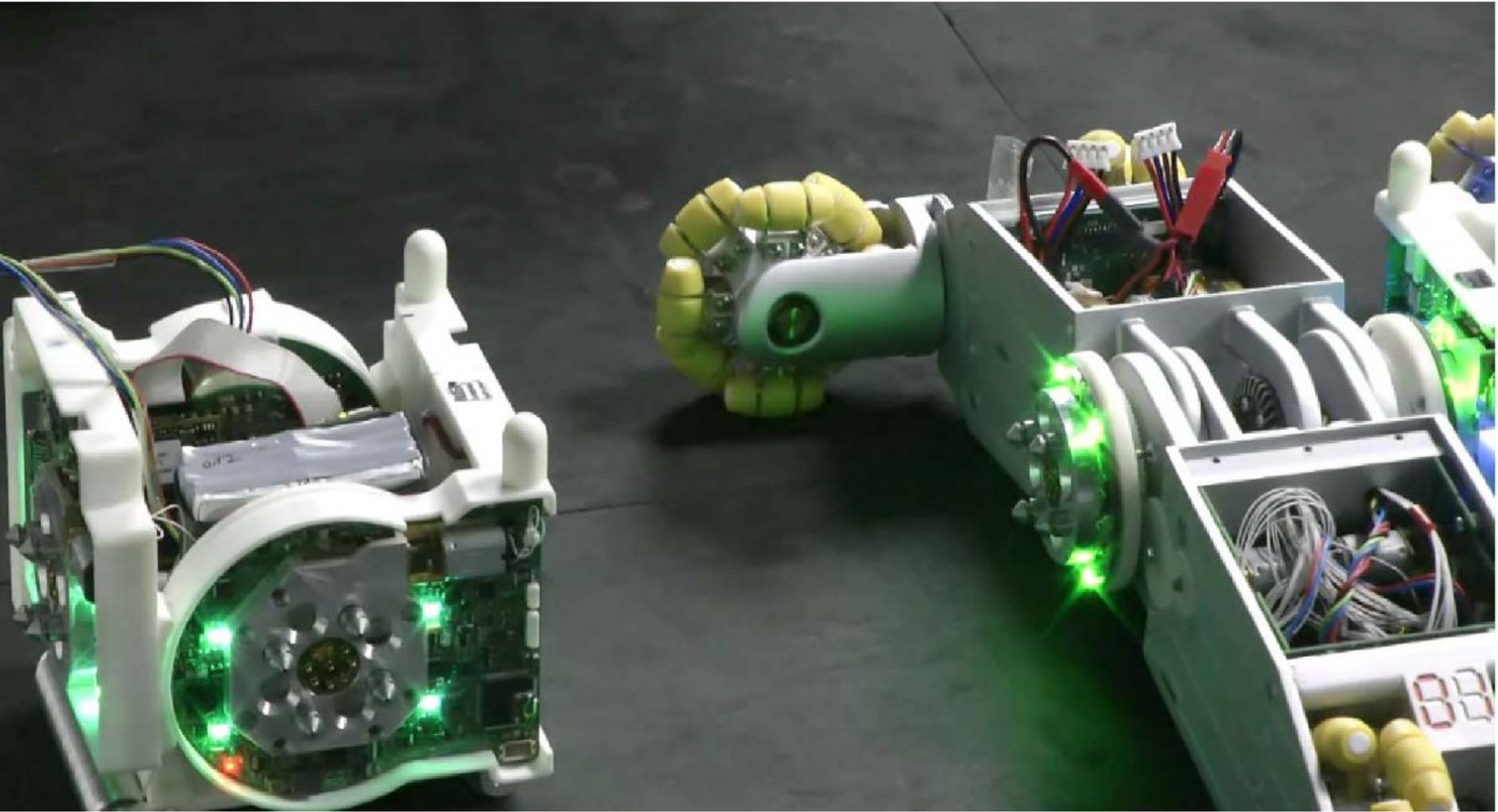}}
	\subfloat[]{\includegraphics[width=0.24\textwidth]{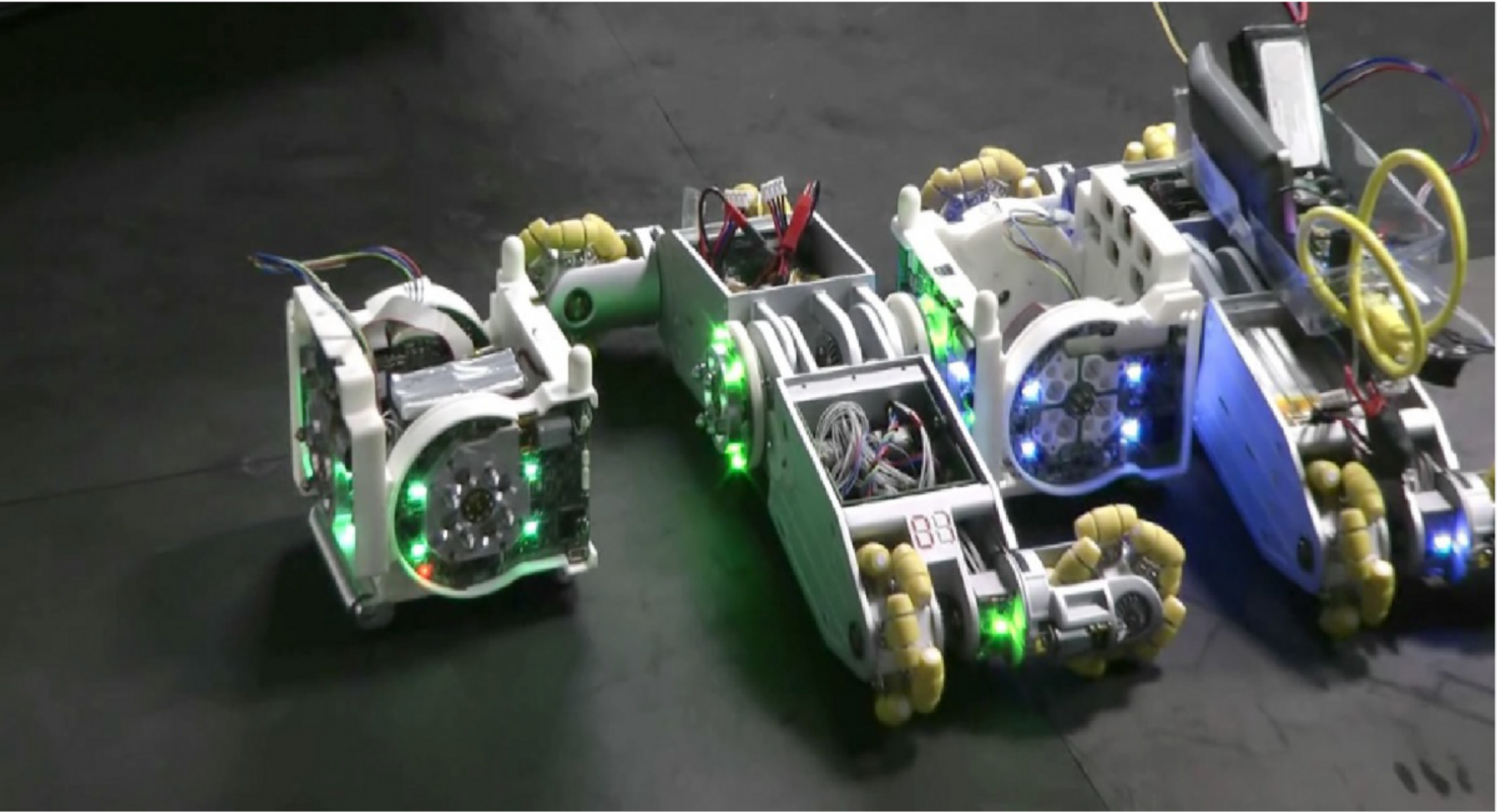}}
	\newline
	\subfloat[]{\includegraphics[width=0.24\textwidth]{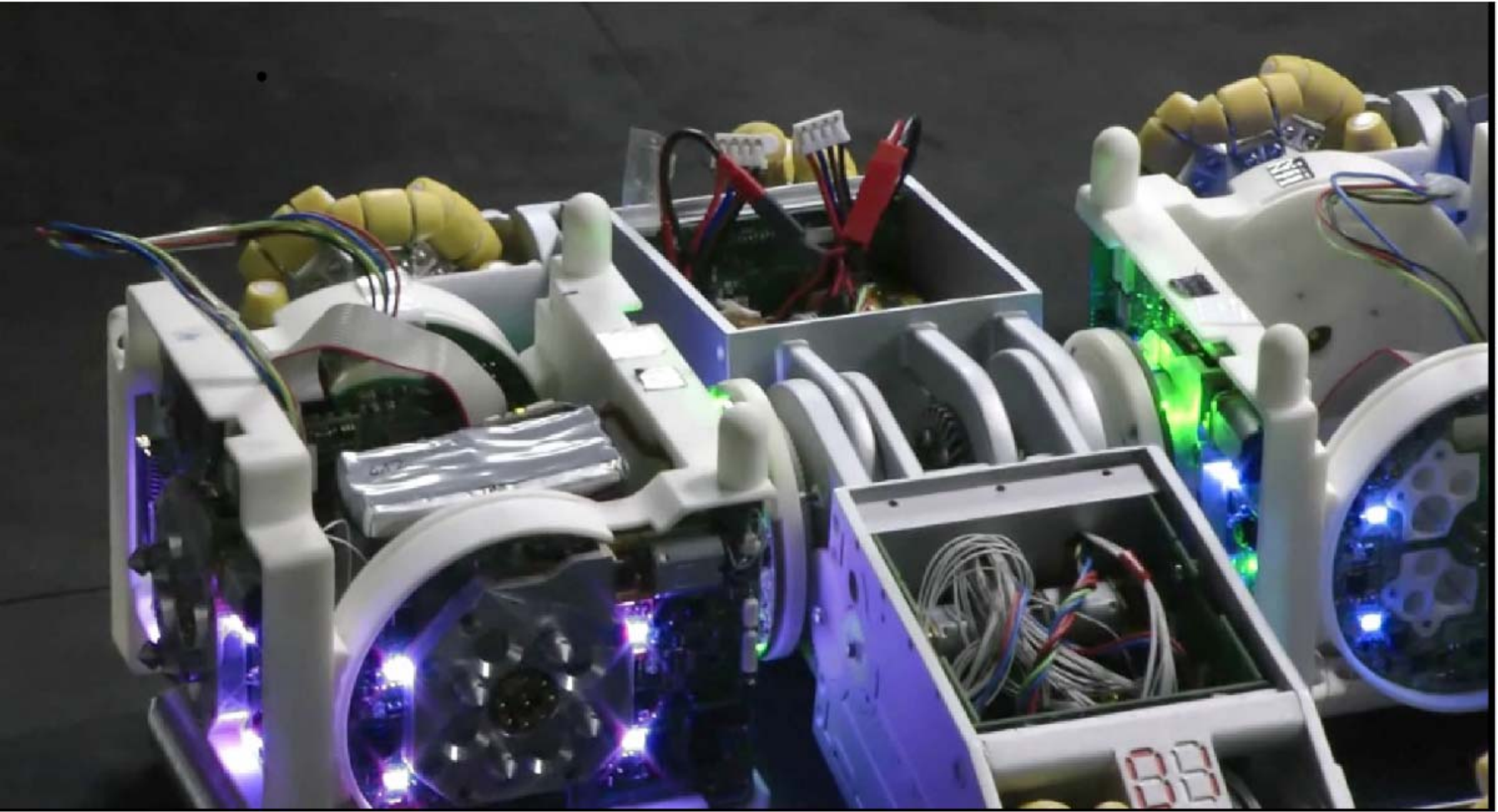}}
	\subfloat[]{\includegraphics[width=0.24\textwidth]{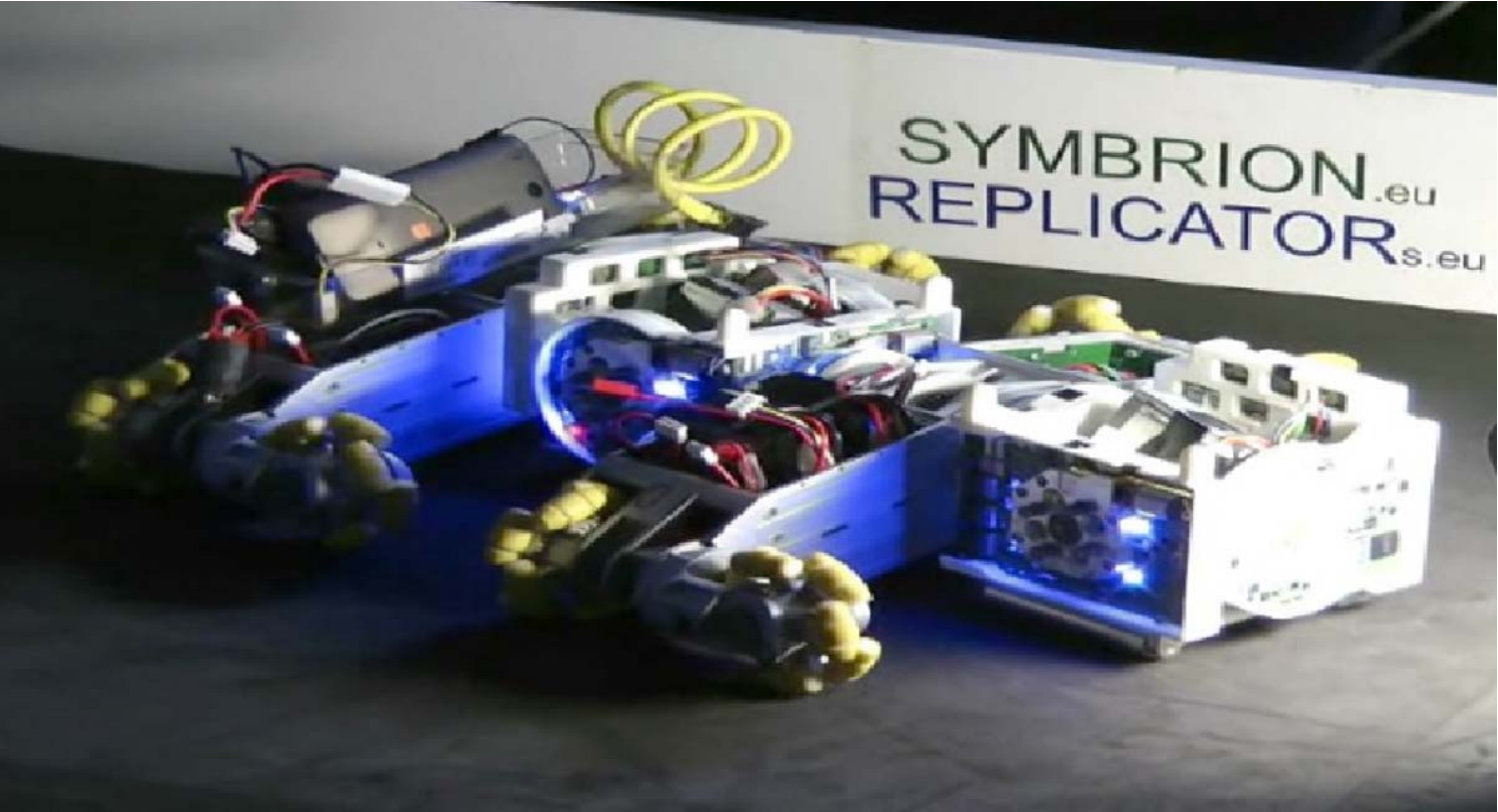}}
	\caption{\small Sequence of the assembling of a heterogeneous organism.}
	\label{fig:organism}
\end{figure}

In this scenario the robots build up a heterogeneous organism of several modules. The objective is to increase the common computational and energetic performance and to create a computationally powerful and efficiently moving organism. By combining two types of robots, the computational power can be raised. Each of the modules adds additional computational power of around 3000~MIPS to the robot organism. Since the Active Wheels can't dock to each other due to their shape, the Backbone robot is used as a connector or skeleton module. The final configuration represents a distributed 4x-2core Blackfin computational systems with an Ethernet bus between CPU nodes. In addition, the Backbone robot adds additional power sources of 33~Wh and sensors to the organism and improves the sensory awareness on the sides, where the Active Wheel has only limited sensors. Thus, two additional Backbone robots, which are not used for locomotion, provide additional 66~Wh energy for the organism. Furthermore, the locomotion of these four robots is faster (31~cm/s for all robots instead of minimal 6~cm/s) and more efficient. Using the wheels of the Active Wheel instead of the screws and lifting up some robots from the ground minimizes the power consumption for locomotion. Thus both robot types can complement each other. The video sequence in Fig.~\ref{fig:organism} shows how multiple robots connect to a heterogeneous organism. In (a)-(b), an Active Wheel connects to two already connected robots. In (c)-(e) the fourth robot docks to the organism and (f) shows the final organism.

\subsection{Improving Reliability: Rescue Scenario}

\begin{figure}[h!]
	\centering
	\subfloat[]{\includegraphics[width=0.165\textwidth]{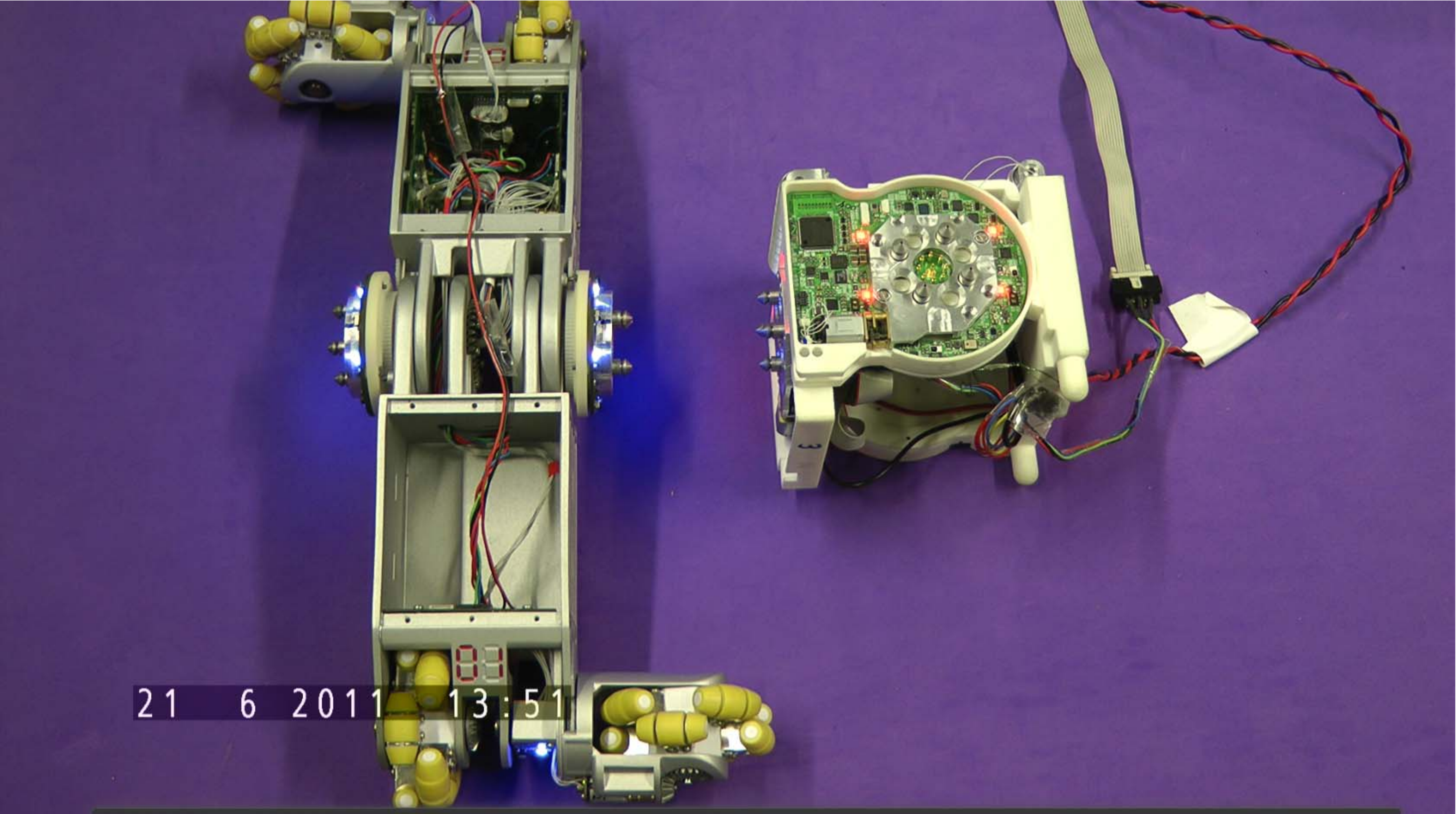}}
   	\subfloat[]{\includegraphics[width=0.165\textwidth]{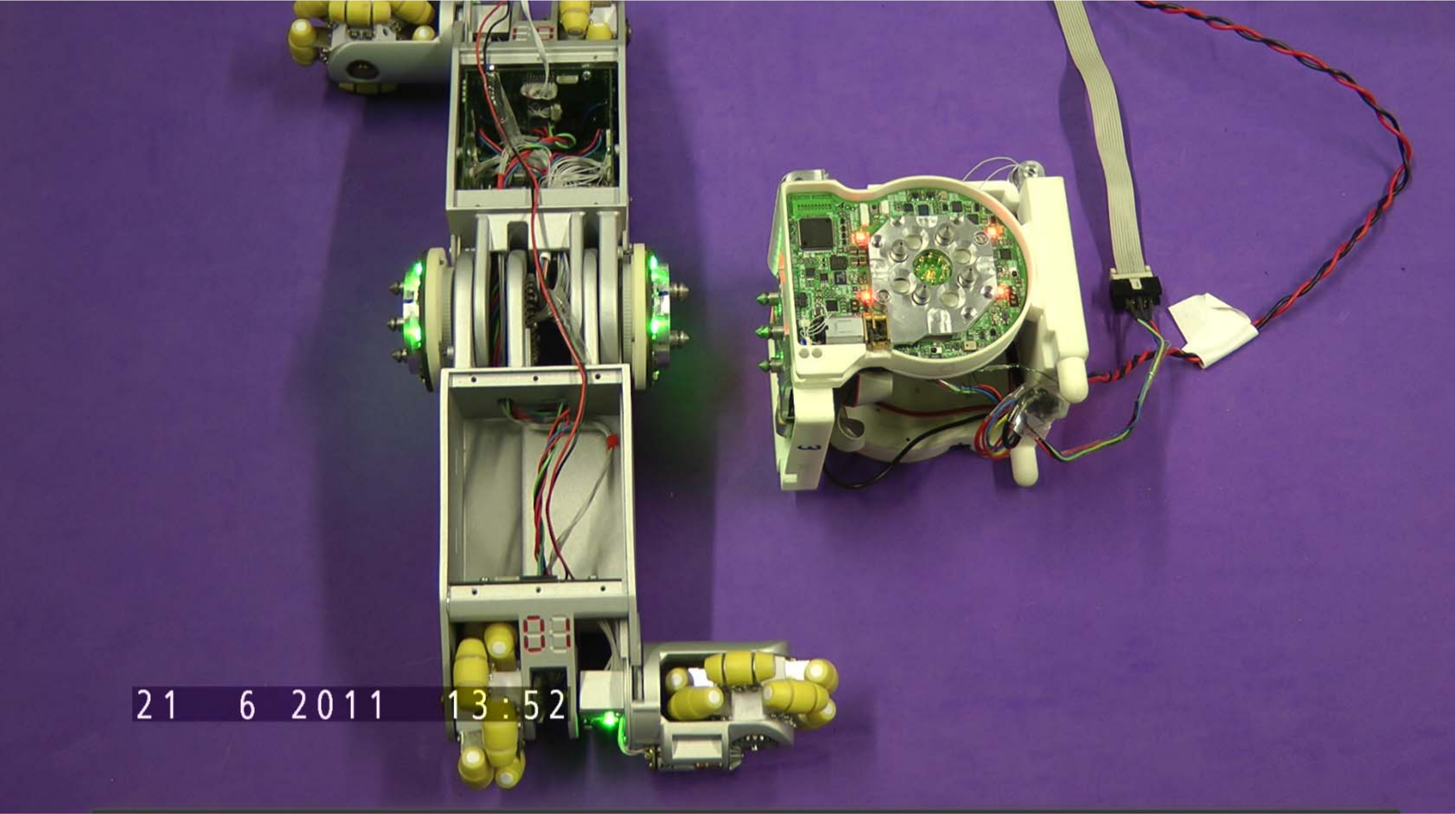}}
    	\subfloat[]{\includegraphics[width=0.165\textwidth]{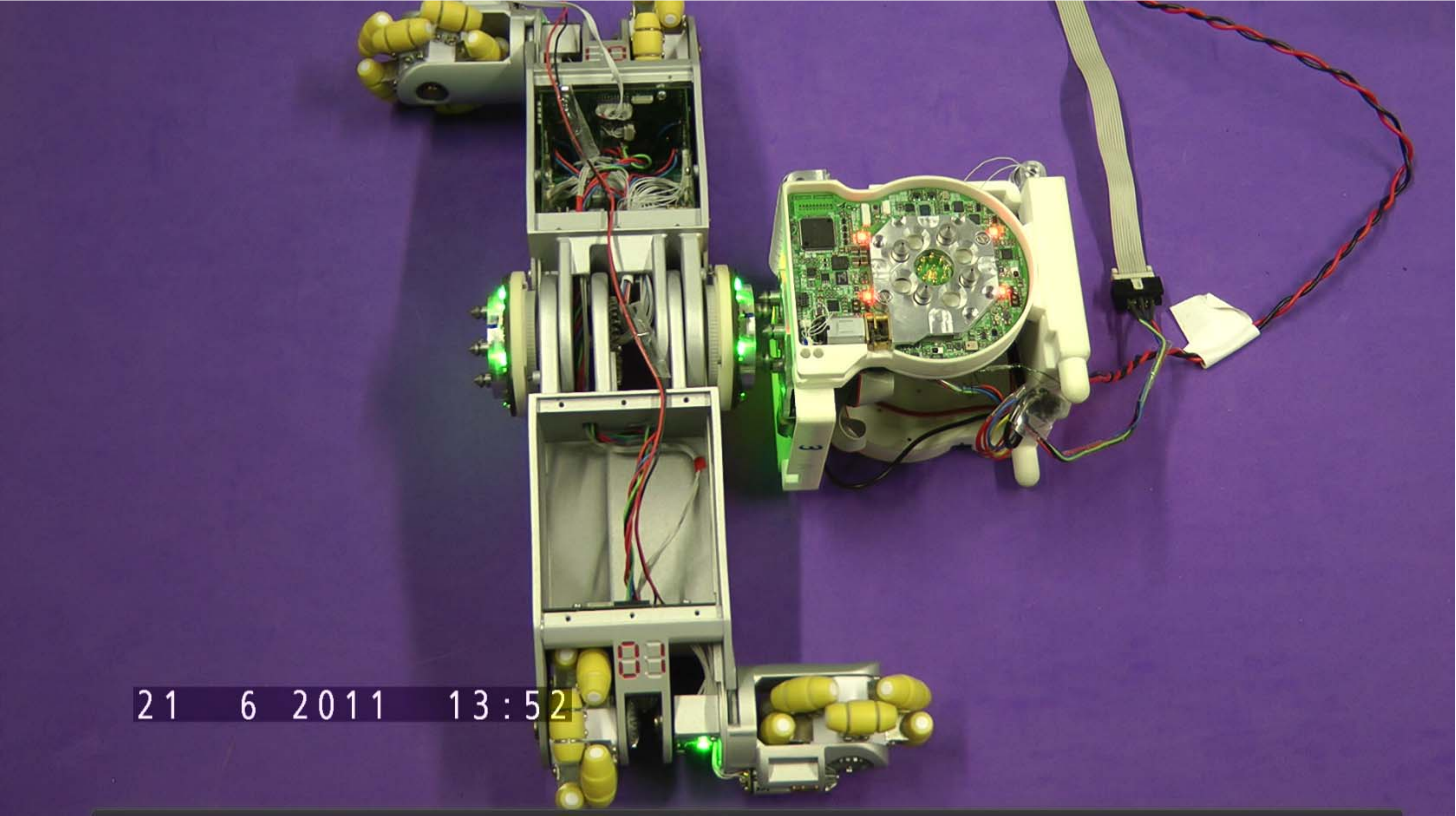}}
	\newline
    	\subfloat[]{\includegraphics[width=0.165\textwidth]{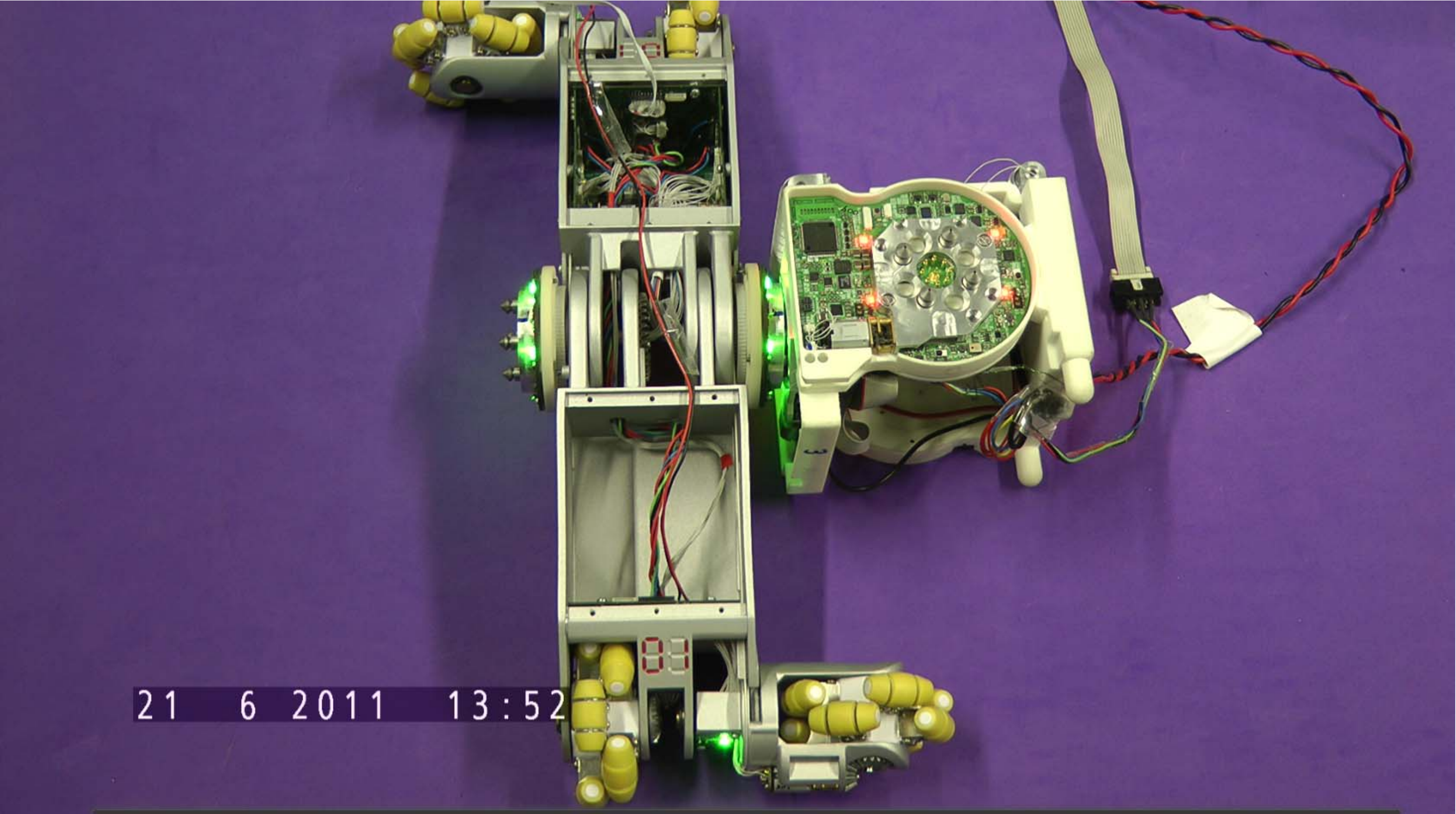}}
    	\subfloat[]{\includegraphics[width=0.165\textwidth]{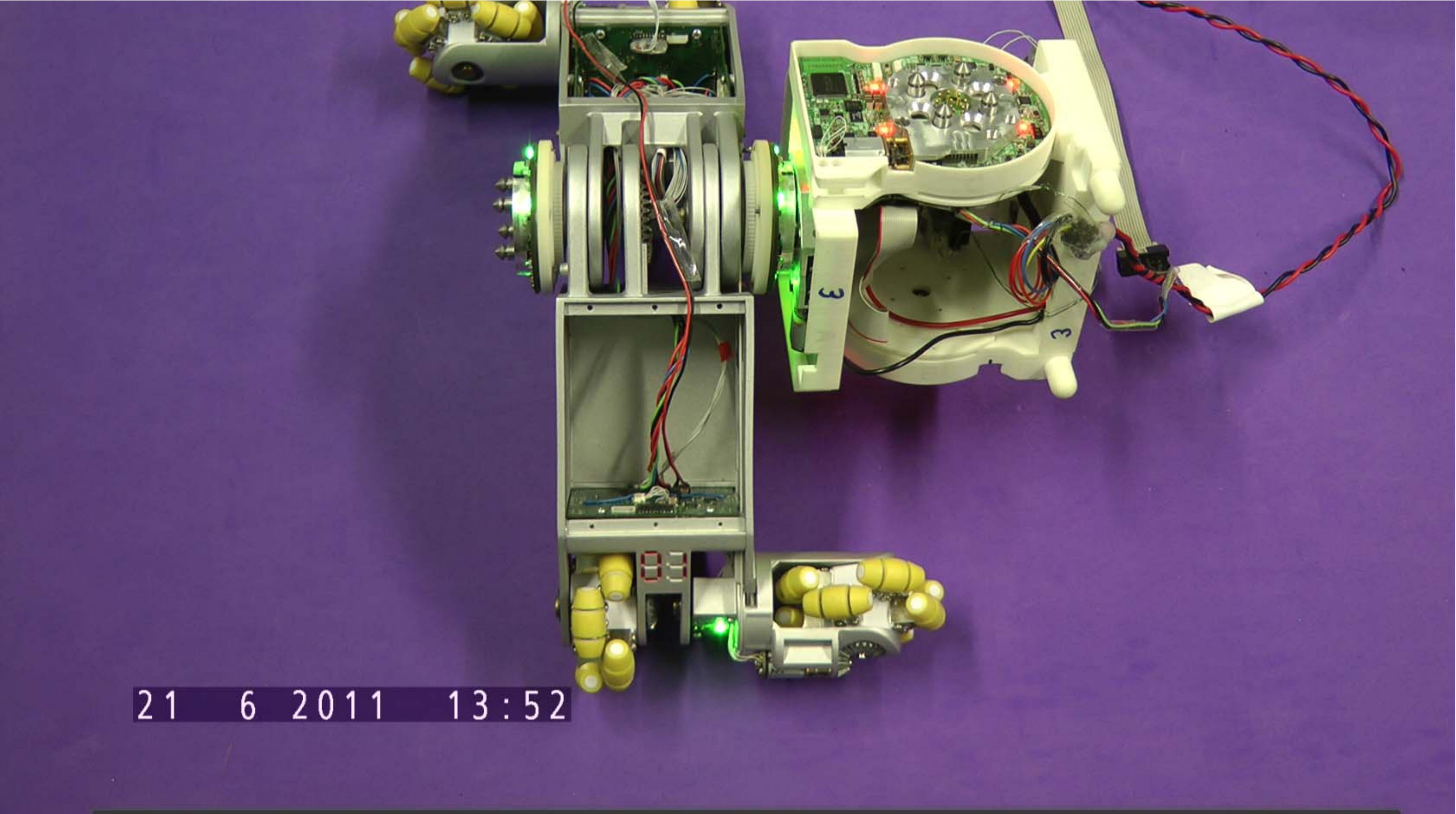}}
    	\subfloat[]{\includegraphics[width=0.165\textwidth]{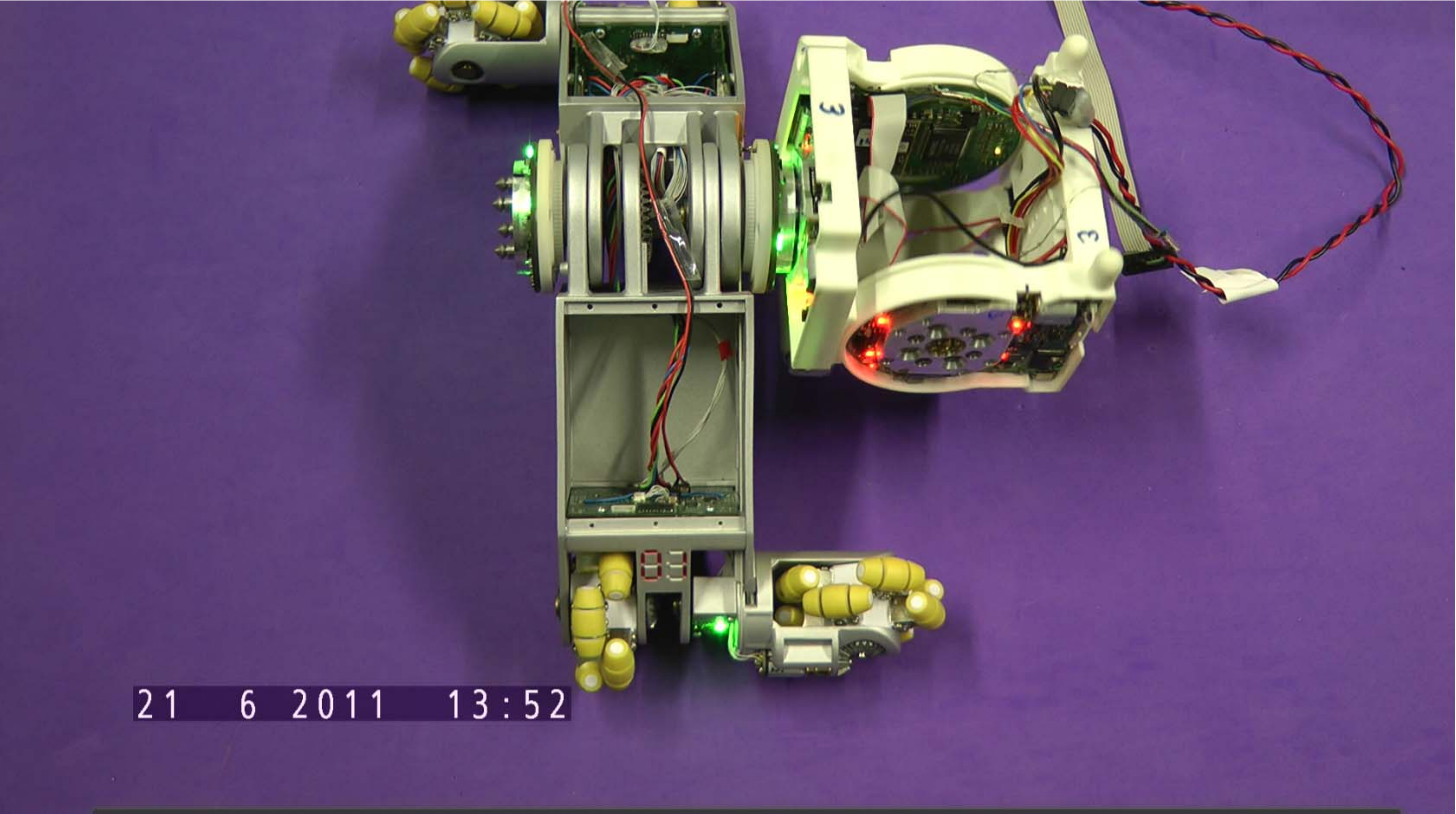}}
	\newline    	
	\subfloat[]{\includegraphics[width=0.165\textwidth]{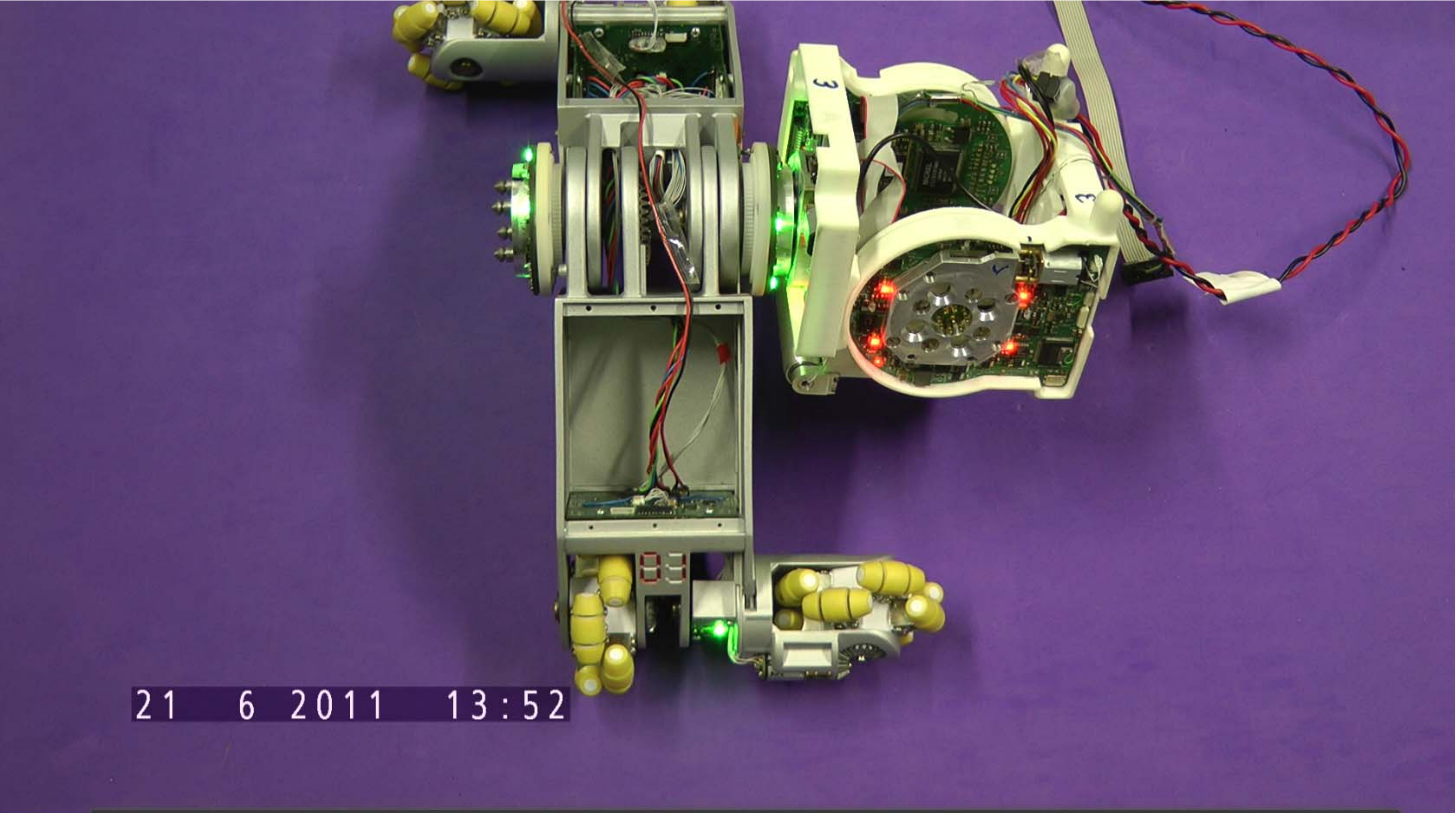}}
    	\subfloat[]{\includegraphics[width=0.165\textwidth]{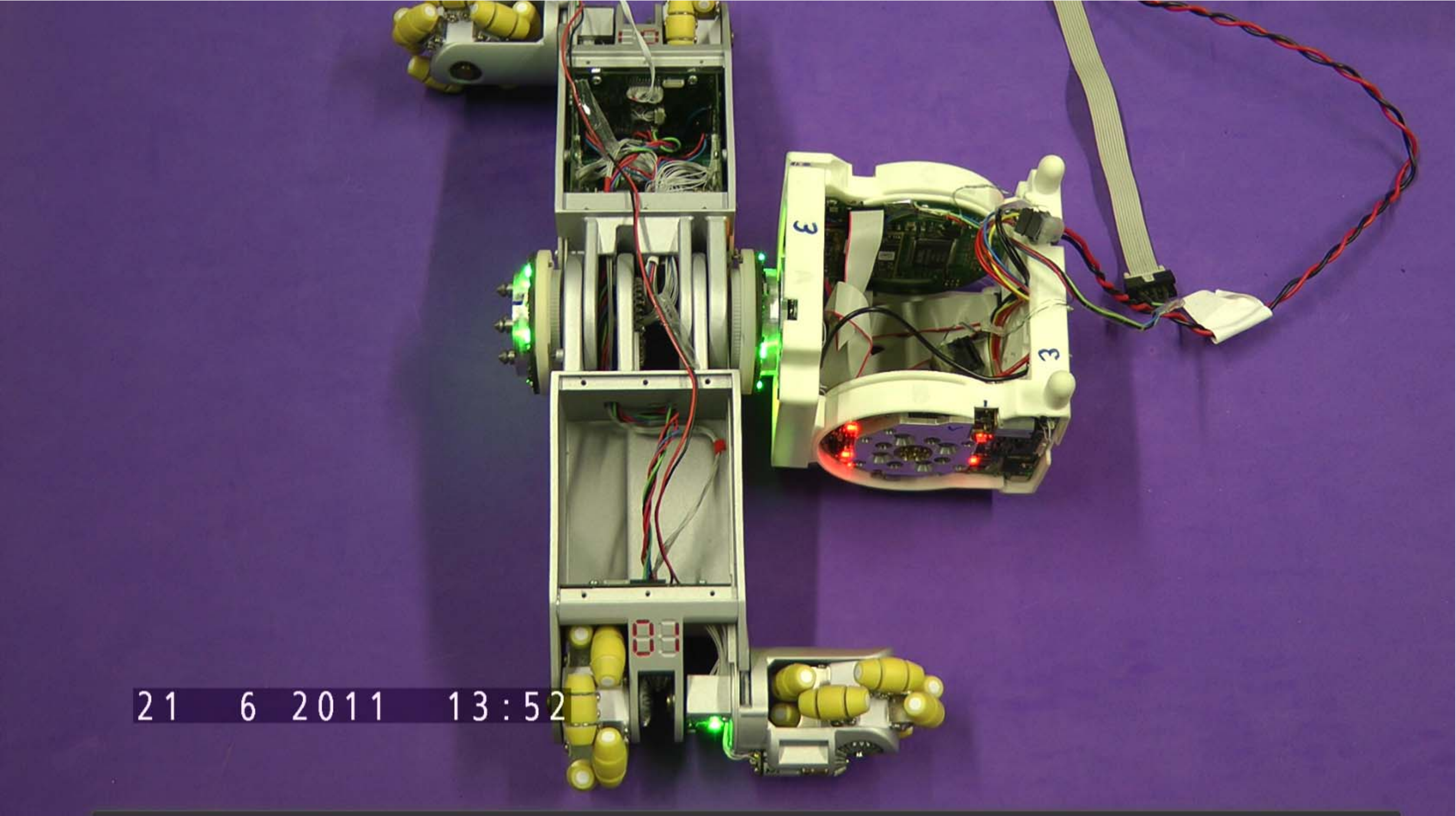}}
    	\subfloat[]{\includegraphics[width=0.165\textwidth]{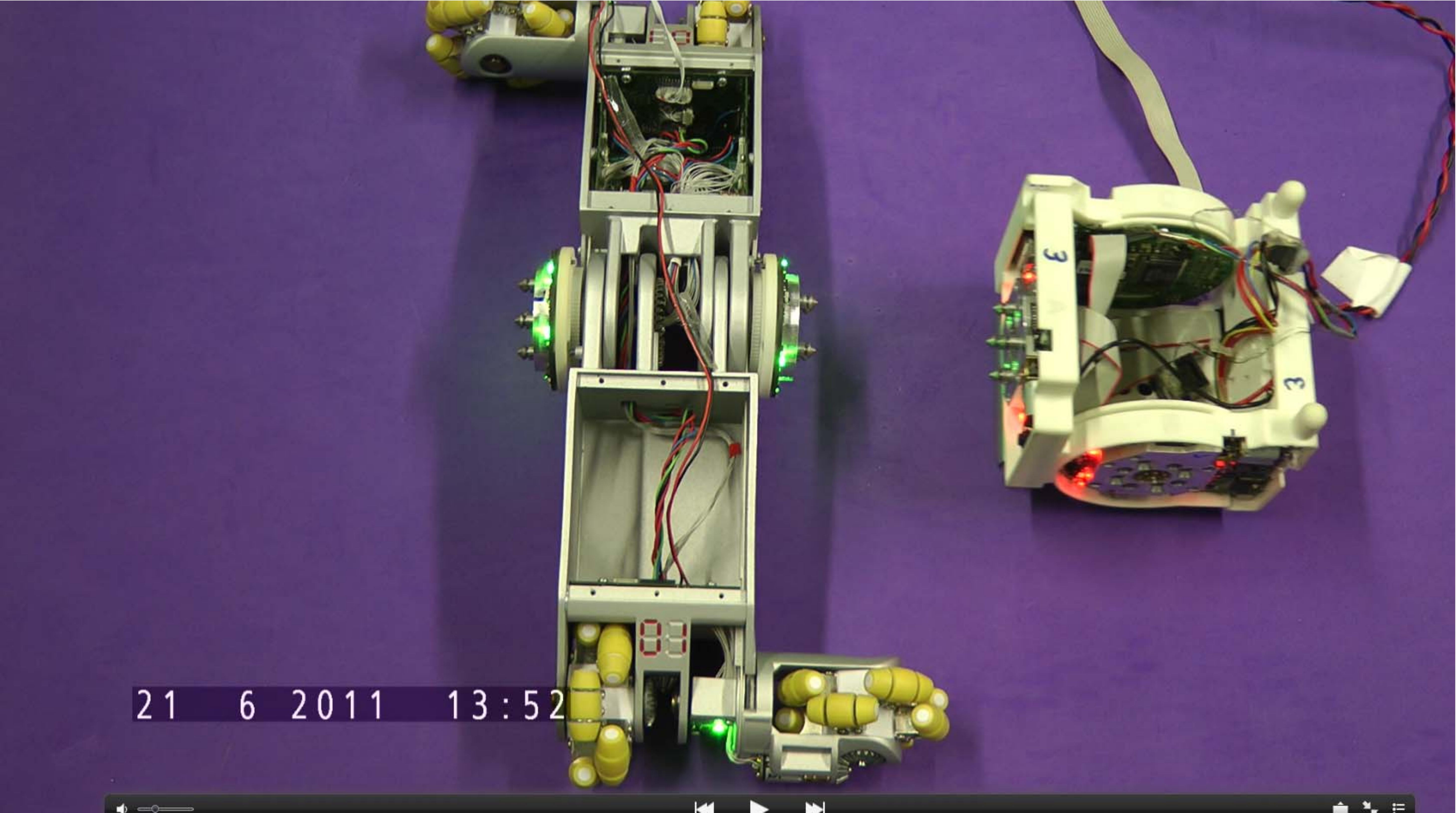}}
	\caption{\small Sequence of the rescue scenario.}
	\label{fig:rescue}
\end{figure}

In this scenario a Backbone robot has fallen on its side for some reason and cannot move any more. This case represents a failure of a platform and excludes it from all further activities. To escape this situation, the robot can call other robots for help. An Active Wheel may catch the cry for help and come to assist the Backbone robot. Because of the rotational docking element on the Active Wheel, it has the capability of rotating another robot. Since the robot cannot be rotated on the ground, the second capability of lifting a robot comes into play. Combining these two features of the Active Wheel allows to lift up the Backbone Robot, turn it around and put it back on its screw drive. A series of images of one of these trials can be seen in Fig.~\ref{fig:rescue}. In (a)-(c) the Active Wheel is approaching the Backbone robot. In (d) it lifts the fallen robot and rotates it in (e)-(g). (h) shows the Active Wheel putting the Backbone robot back onto the ground and finally in (i) the Backbone robot can continue its activities.

\section{Conclusion}
\label{sec:conclusion}

In this work we presented the heterogeneous approach of a self-reconfigurable system, its design rationales, main argumentations and challenges. Four heterogeneous platforms, specialized in different tasks, as well as their common elements such as docking mechanism or electronics were shortly demonstrated.

Originally, the self-reconfigurable platform was planned as a homogeneous swarm-like robot of the size $\rm\sim 50~cm^3$, weighing around 200~g. Multi-robot organisms should consist of a large number of such modules. The main factors which introduced and continuously increased the level of heterogeneity are physical constraints imposed on the system in the organism mode: the number of modules to be lifted, mechanical strength of docking elements or requirements for energy/communication buses. This finally led to the size of $\rm\sim 1000~cm^3$ and the weight of around 1~kg per module and a lower number of modules in the artificial organism.

Generalizing this experience, we argue that the degree of heterogeneity is defined by a diversity of required functionalities. As demonstrated by the ATRON and MTRAN systems, morphing capabilities can be achieved with only one type of modules. As soon as any additional requirement is introduced, the optimization of the system's performance leads to an appearance of specialized modules. The optimal degree of heterogeneity is a trade-off between optimization of performance and such factors
like "replaceability of modules", "costs of manufacturing", "maintenance", "available developmental power", i.e. with organizational/managemental factors. 

Heterogeneity strongly requires several universal elements, this can be denoted as "homogeneity inside of heterogeneity". In our example, these are docking mechanism, communication and energy sharing system. To some extent, compatible electronic and software frameworks in all modules are also necessary to minimize the effort of development and for performing experiments.

We demonstrated the benefit of a heterogeneous system by two qualitative experiments: aggregation into an artificial organism, where the organism became much more computationally and energetically powerful than a sum of non-aggregated modules, and by a rescue scenario, where heterogeneous robots assist each other in increasing the reliability of the whole system.

\section*{Acknowledgement}

The authors are supported by the following grants: EU-ICT FET project "SYMBRION", grant agreement no. 216342; EU-ICT project "REPLICATOR", grant agreement no. 216240. Additionally, we want to thank all members of the projects for their cooperation and fruitful discussions.

\small


\end{document}